\documentclass[11pt]{article}
\usepackage[final]{acl}

\usepackage{times}
\usepackage{latexsym}
\usepackage[T1]{fontenc}
\usepackage[utf8]{inputenc}
\usepackage{microtype}
\usepackage{inconsolata}
\usepackage{graphicx}

\usepackage{tcolorbox}
\usepackage{fontawesome5}
\usepackage{xspace}  
\usepackage{xcolor}  
\usepackage[table,xcdraw]{xcolor}
\usepackage{multicol}
\usepackage{multirow}
\usepackage{booktabs}
\usepackage{pifont}
\usepackage[table]{xcolor}
\definecolor{avggray}{gray}{0.93}
\usepackage{amssymb}
\usepackage{amsmath} 

\newcommand{\cmark}{\ding{51}} 
\newcommand{\xmark}{\ding{55}} 
\definecolor{lightred}{RGB}{255, 230, 230}
\newcommand{\highlightred}[1]{%
  \colorbox{lightred}{\parbox[t]{\linewidth}{\strut #1 \strut}}%
}

\definecolor{mybg}{HTML}{FFFDE9}

\graphicspath{{img/}}   

\title{Rethinking Stepwise Model Routing: A Cost-Efficient Table Reasoning Perspective}
\author{
  \textbf{Shenghao Ye\textnormal{\textsuperscript{1}\footnotemark[1]}},
  \textbf{Yuxiang Wang\textnormal{\textsuperscript{2}\footnotemark[1]}},
  \textbf{Yu Guo\textnormal{\textsuperscript{1}\footnotemark[1]}},
  \textbf{Tao Zhang\textnormal{\textsuperscript{1}}},
  \textbf{Shuangwu Chen\textnormal{\textsuperscript{1}\footnotemark[2]}}, \\
  \textbf{Zijian Wen\textnormal{\textsuperscript{1}}},
  \textbf{Qihui Zhu\textnormal{\textsuperscript{1}}},
  \textbf{Mengjie Zhang\textnormal{\textsuperscript{1}}},
  \textbf{Dong Jin\textnormal{\textsuperscript{4}}},
  \textbf{Wang Luo\textnormal{\textsuperscript{3}}},
  \textbf{Jian Yang\textnormal{\textsuperscript{1}}}
\\
  \textsuperscript{1}University of Science and Technology of China 
\\
  \textsuperscript{2} The University of Melbourne \textsuperscript{3} Sun Yat-sen University
\\
  \textsuperscript{4}Institute of Artificial Intelligence, Hefei Comprehensive National Science Center
\\
  \texttt{\{ssh0321y, yukariguo\}@mail.ustc.edu.cn}
\\
    \texttt{yuxiang.wang8@student.unimelb.edu.au}
}

\begin{document}
\maketitle
\renewcommand{\thefootnote}{\fnsymbol{footnote}}
\footnotetext[1]{Equal contribution}
\footnotetext[2]{Corresponding authors}
\begin{abstract}
Large Reasoning Models (LRMs) achieve strong performance on table reasoning tasks but incur substantial inference cost due to long reasoning traces. Stepwise model routing mitigates this issue by dynamically assigning reasoning steps to smaller or larger models. However, stepwise model routing for table reasoning remains underexplored. Through empirical analysis, we find that reasoning steps involving tables contain two types of tokens with distinct uncertainty distributions: table tokens grounded in table structure, such as cell values and headers, and text tokens representing surrounding natural-language reasoning. The uncertainty of both token types is correlated with the risk that the model makes an error in the next reasoning step. However, existing methods fail to model them separately, leading to suboptimal routing decisions. To address this, we propose EcoTab, a table-aware stepwise routing framework for efficient table reasoning. At each reasoning step, EcoTab separately estimates the uncertainties of table tokens and text tokens, maps them to next-step failure risks for the small model, and combines the two risks for routing. Experiments on multiple table reasoning benchmarks show that EcoTab consistently outperforms strong baselines and achieves a better balance between accuracy and efficiency.
\end{abstract}

\section{Introduction}


Table reasoning plays a critical role in real-world applications, including data analytics \citep{zhao2024docmath}, fact verification \citep{parikh2020totto} and scientific reporting \citep{newman2024arxivdigestables}.
However, it remains challenging because tables contain complex structures and implicit relations across rows and columns. Recent Large Reasoning Models (LRMs), such as DeepSeek-R1 \citep{guo2025deepseek} and OpenAI's o-series \citep{pfister2025understanding}, improve performance by using test-time scaling to produce long reasoning chains during inference. Despite their strong results, this process introduces high computational overhead. The large model size and heavy token usage make LRMs difficult to deploy for table reasoning in latency-sensitive and resource-constrained settings \citep{zeng2026glimprouter}.

To mitigate this bottleneck, stepwise model routing \citep{shi-etal-2025-speccot,lee2026confidence} has emerged as a promising direction. It decomposes the inference process into multiple reasoning steps, allocating simpler steps to smaller, cheaper models and more complex ones to larger, more expensive models. In this way, stepwise model routing offers an effective balance between efficiency and performance \citep{fernandez2026radar}. Existing methods, such as SpecCoT \citep{shi-etal-2025-speccot} and SpecReason \citep{pan2025specreason}, perform well on free-form text reasoning tasks like mathematical reasoning. However, their effectiveness on structured table reasoning tasks remains underexplored. 


%

\begin{figure*}[t]
    \centering
    \includegraphics[width=0.98\linewidth]{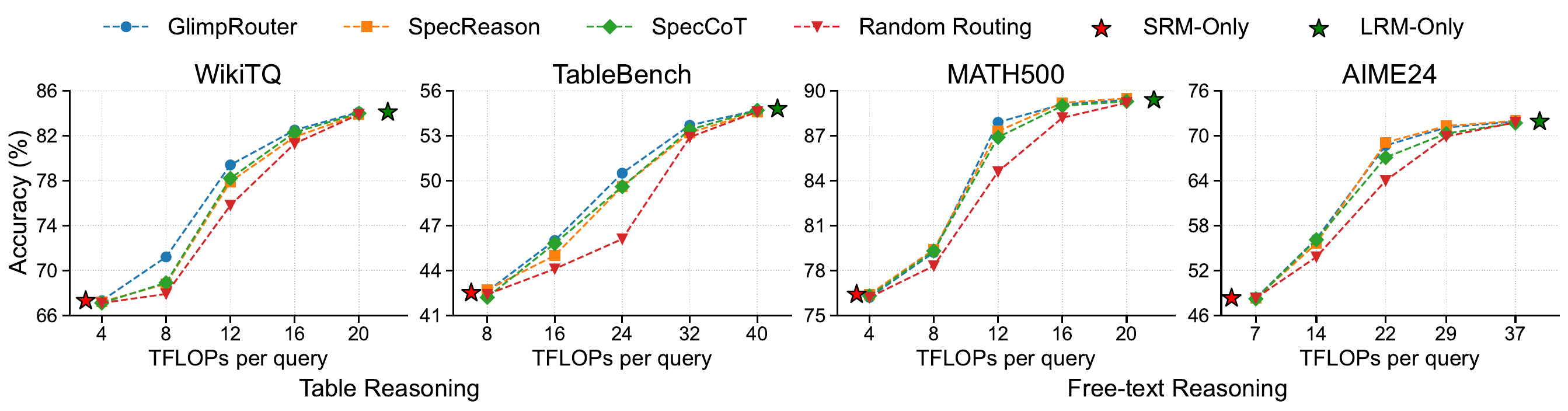}
    \caption{Effectiveness analysis on table and free-form text reasoning tasks. WikiTQ and TableBench represent table reasoning benchmarks, while MATH500 and AIME24 correspond to free-form text reasoning benchmarks.}
    \label{fig:empirical_fig1}
    \vspace{-12pt}
\end{figure*}

To examine this question, we revisit several stepwise model routing methods and evaluate their effectiveness on table reasoning tasks, as detailed in Sec.~\ref{sec2}. Our analysis reveals a clear gap in the efficiency and performance trade-off between free-form text reasoning and table reasoning. We find that existing methods often misroute ``table-specific steps'', such as retrieving the relevant subtable for a question or performing numerical operations over tabular content. To understand this failure, we further analyze the root cause and identify a key insight: such steps contain two types of tokens with distinct uncertainty distributions, namely table tokens, grounded in table structure such as cell values and headers, and text tokens, which reflect the surrounding natural language reasoning. We further find that the uncertainty of both token types is correlated with the next-step failure risk of the small model, which in turn is informative for model selection. However, existing methods lack joint modeling of table tokens and text tokens, resulting in poor routing performance on table reasoning tasks.

Motivated by these findings, we propose \textbf{EcoTab}, an \textbf{\underline{e}}ffi\textbf{\underline{c}}ient table-aware stepwise model r\textbf{\underline{o}}uting framework for \textbf{\underline{tab}}le reasoning. EcoTab is built on a simple intuition: table tokens and text tokens exhibit different uncertainty distributions and should therefore be modeled separately during routing. For each reasoning step, EcoTab first identifies table tokens and text tokens in the current reasoning step and estimates their uncertainties separately. To account for their distinct distributions, EcoTab constructs two offline risk mappings that convert these uncertainties into next-step failure risks, where each risk reflects the likelihood that the small model will fail on the next step. Finally, EcoTab combines the two failure risks into a unified routing score and compares it with a threshold to decide whether the next step should be generated by a small model or a large one.


\textbf{Our Contributions}. 
\underline{(1) \textit{New Perspective}.} We conduct the first systematic study of stepwise model routing for table reasoning. We reveal that table reasoning steps contain two types of tokens with distinct uncertainty distributions, explaining why existing routing methods designed for free-form text fail on table reasoning. \underline{(2) \textit{Novel Framework}.} We propose EcoTab, an efficient table-aware stepwise model routing framework for table reasoning. \underline{(3) \textit{SOTA Performance}.} Experiments on multiple table reasoning benchmarks show that EcoTab consistently outperforms strong baselines and achieves a better balance between accuracy and efficiency.

\vspace{-2pt}

\section{Preliminary}
\vspace{-2pt}

\paragraph{Table Reasoning with LRMs.}
Given a table $T$ and a natural language query $Q$, an LRM generates a sequence of reasoning steps $s_1, \dots, s_n$, denoted as $s_{1:n}$, where each step $s_i$ contains $k_i$ tokens. Following prior studies~\citep{pan2025specreason,zeng2026glimprouter}, we segment reasoning traces into steps using the newline delimiter ``\texttt{\textbackslash n\textbackslash n}''.

\vspace{-4pt}

\paragraph{Stepwise Model Routing.}
Given the current reasoning prefix $s_{1:n}$, stepwise model routing dynamically chooses between the small reasoning model (SRM) and the large reasoning model (LRM) to generate the next reasoning step $s_{i+1}$, so as to improve computational efficiency. Formally, the next step is generated as
\begin{equation}
s_{i+1} \sim p_{\theta_{i+1}}(\cdot \mid T, Q, s_{1:i}), 
\theta_{i+1} = r(\mathcal{I}_{i+1})
\end{equation}
where $p_{\theta_{i+1}}$ denotes the probability distribution of the selected reasoning model, and $\theta_{i+1} \in \{\theta_M, \theta_m\}$ corresponds to the LRM or the SRM. Here, $r(\cdot)$ denotes the routing function, and $\mathcal{I}_{i+1}$ denotes the routing information used to determine the model for step $i+1$. Depending on the routing function, $\mathcal{I}_{i+1}$ may come from the previously generated step $s_i$~\citep{lee2026confidence}, or from a lightweight preview or draft of the next step~\citep{pan2025specreason,zeng2026glimprouter,shi-etal-2025-speccot}. In this work, we use $s_i$ as $\mathcal{I}_{i+1}$, which avoids additional token generation overhead.

\begin{figure}[t]
    \centering
    \includegraphics[width=0.8\linewidth]{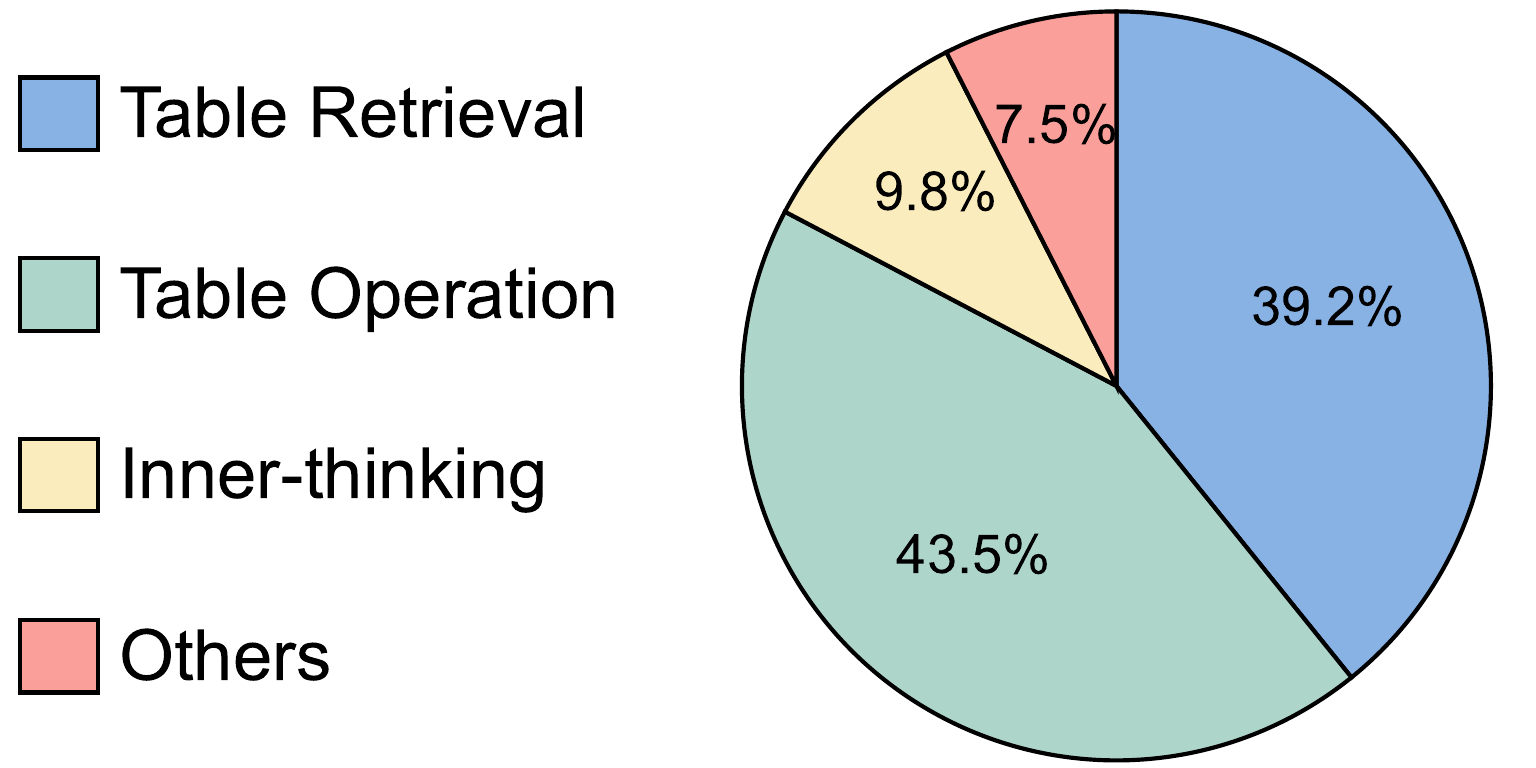}
    \caption{Error distribution across four step categories over 1000 incorrect cases under GlimpRouter.}
    \label{fig:empirical_fig2}
    \vspace{-12pt}
\end{figure}

\vspace{-4pt}

\section{Motivation}
\label{sec2}
\vspace{-2pt}
In this section, we investigate why table reasoning requires a dedicated stepwise model routing beyond existing methods designed for free-form text reasoning. This leads to our first research question:

\begin{tcolorbox}[
    colback=yellow!10!white,      %
    colframe=black,            %
    coltitle=white,            %
    fonttitle=\bfseries,       %
    boxrule=.7pt,
    width=\linewidth,
    top=1mm,
    bottom=1mm,
    left=2mm,                  %
    right=2mm,                 %
    before skip=6pt, after skip=6pt
]
\vspace{-1pt}
\textit{\textbf{RQ1 --}} Do free-form text routing methods adapt effectively to table reasoning?
\vspace{-1pt}
\end{tcolorbox}

\textbf{Effectiveness Analysis.} 
The effectiveness of stepwise model routing depends on reaching LRM-only accuracy at lower cost, which we measure by FLOPs. We evaluate several representative methods, including GlimpRouter \citep{zeng2026glimprouter}, SpecReason \citep{pan2025specreason}, SpecCoT \citep{shi-etal-2025-speccot}, and Random Routing. We adopt Qwen3-1.7B and Qwen3-14B \citep{yang2025qwen3} as the SRM and LRM. The evaluation covers table reasoning benchmarks, including WikiTQ \citep{pasupat-liang-2015-compositional} and TableBench \citep{wu2025tablebench}, as well as free-form text reasoning benchmarks such as MATH500 \citep{lightman2023letsverifystepstep} and AIME24. More experimental details are provided in Appendix~\ref{sec:appendix1}.
As shown in Figure~\ref{fig:empirical_fig1}, assigning more steps to the stronger LRM naturally increases both FLOPs and accuracy. However, existing routing methods are much less efficient on table reasoning than on free-form text reasoning. On free-form text benchmarks, they achieve near-LRM performance with only about 60\% of the full FLOPs. In contrast, on table reasoning benchmarks, they require nearly 80\% of the full FLOPs to reach a similar level of performance. This efficiency gap indicates that existing methods do not transfer effectively to table reasoning.

\begin{figure}[t]
    \centering
    \includegraphics[width=0.93\linewidth]{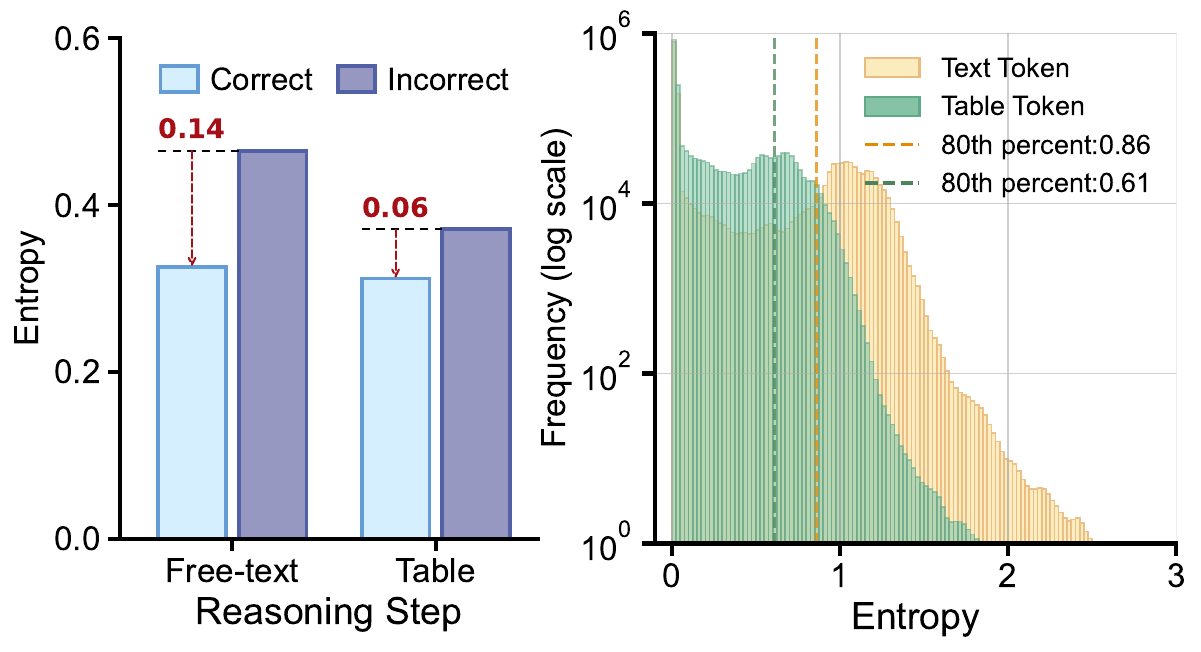}
    \caption{(Left) Difference in average entropy between correct and incorrect steps for Free-Text steps and Table-specific steps. (Right) Entropy distributions of table tokens and text tokens within Table-specific steps.}
    \label{fig:empirical_fig3}
    \vspace{-14pt}
\end{figure}

\textbf{Error Analysis}. To understand the source of this efficiency gap, an error analysis of the routing process is conducted. Specifically, we randomly sample 1000 erroneous cases that are correctly solved under the LRM-only setting but fail under GlimpRouter \citep{zeng2026glimprouter}, and ask human experts to identify the failure step in each case and classify it into one of four error types, following TaTToo \citep{zou2026tattoo}: (i) \textit{Table Retrieval}, (ii) \textit{Table Operation}, (iii) \textit{Inner-Thinking}, and (iv) \textit{Others} (defined in Appendix \ref{sec:appendix1}). \textit{Table Retrieval} and \textit{Table Operation} are defined as ``table-specific steps'', as they are unique to table reasoning, while \textit{Inner-Thinking} and \textit{Others} are regarded as ``free-text steps''.
As shown in Figure~\ref{fig:empirical_fig2}, 82.7\% of routing errors arise from table-specific steps. This indicates that free-form text routing methods fail to properly route these steps, often assigning them to the SRM when the LRM is actually needed.
\newline
\textbf{Finding of RQ1}. Free-form text routing methods fail to properly route table-specific steps, leading to a notable efficiency gap in table reasoning.


\begin{figure}[t]
    \centering
    \includegraphics[width=1.035\linewidth]{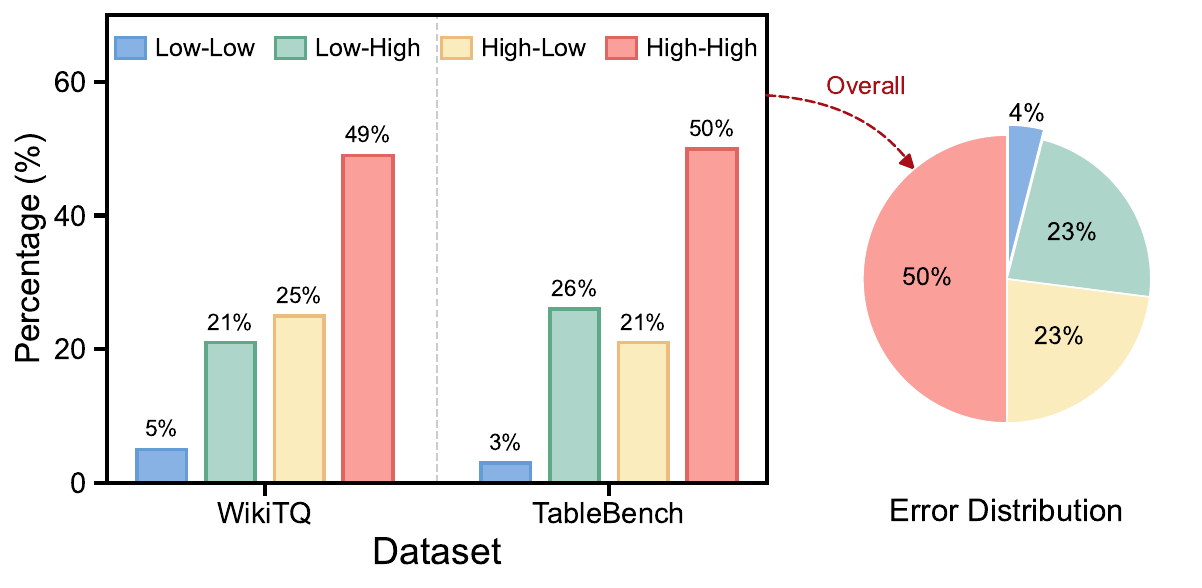}
    \caption{(Left) Error percentages across four entropy groups on WikiTQ and TableBench for Table-specific steps. (Right) Overall error distribution across the four groups.}
    \label{fig:empirical_fig6}
    \vspace{-14pt}
\end{figure}

It raises a subsequent research question:
\vspace{-3pt}
\begin{tcolorbox}[
    colback=yellow!10!white,      %
    colframe=black,            %
    coltitle=white,            %
    fonttitle=\bfseries,       %
    boxrule=.7pt,
    width=\linewidth,
    top=1mm,
    bottom=1mm,
    left=2mm,                  %
    right=2mm,                 %
    before skip=6pt, after skip=6pt
]
\vspace{-1pt}
\textit{\textbf{RQ2 --}} Why do free-form text routing methods fail on table-specific reasoning steps?
\vspace{-1pt}
\end{tcolorbox}

\textbf{Table Tokens Differ from Text Tokens. }
To understand this failure, we compare free-text steps with table-specific steps. Following GlimpRouter, we use the average step entropy as the routing score and randomly sample 500 correct steps and 500 incorrect steps. As shown in Figure~\ref{fig:empirical_fig3} (left), free-text steps show a clear separation between correct and incorrect cases, with an average entropy gap of 0.14. In contrast, the gap for table-specific steps drops sharply to 0.06. This suggests that step-level entropy is much less informative for routing table-specific steps. We then analyze table-specific steps at the token level by separating each step into \textit{table tokens} and \textit{text tokens}. As shown in Figure~\ref{fig:empirical_fig3} (right), the two token types exhibit clearly different entropy distributions. This suggests that they play different roles during reasoning, which is also consistent with prior studies on table reasoning \citep{wang2025needleinatable,zou2026tattoo,li2025table}.

\begin{figure*}[t]
    \centering
    \includegraphics[width=0.96\linewidth]{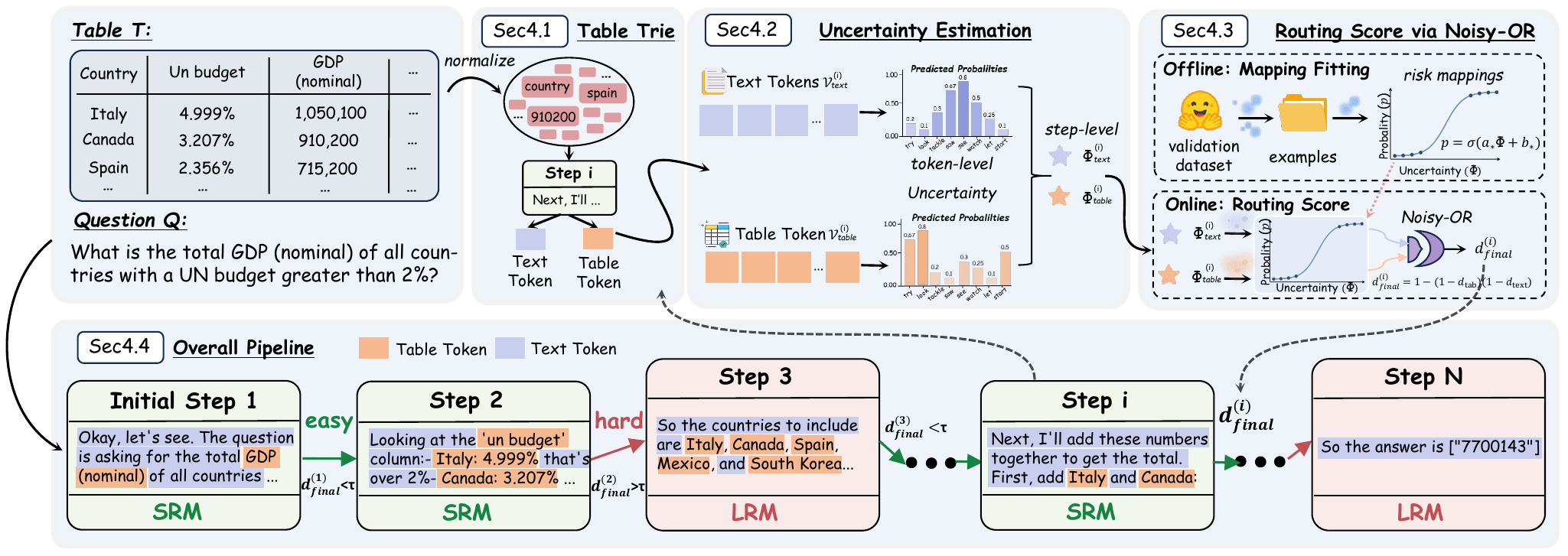}
    \caption{Overview of the EcoTab framework. By separately modeling table tokens and text tokens in each reasoning step $s_i$, EcoTab enables more effective routing between the SRM and LRM for table reasoning.}
    \label{fig:method}
    \vspace{-14pt}
\end{figure*}
 
\textbf{Both Token Types Matter for Routing.} 
Building on the analysis above, we further ask whether both table tokens and text tokens are related to the next-step failure risk. Specifically, we compute the average entropy of table tokens and text tokens for 1,000 sampled steps, including 500 correct steps and 500 incorrect steps. For each type of token, we use the 70th percentile as the threshold \citep{notin2021improving}. A score above the threshold is labeled as High, and a score below it is labeled as Low. This gives four groups, namely High-High, High-Low, Low-High, and Low-Low. We then examine how the 500 incorrect steps are distributed across these four groups. As shown in Figure~\ref{fig:empirical_fig6}, errors are not concentrated only in the High-High group. A substantial portion also falls into the High-Low and Low-High groups. This shows that either high table-token uncertainty or high text-token uncertainty alone can be associated with failure risk. Therefore, effective routing should consider both token types rather than relying on only one of them.

\vspace{-5pt}
\begin{tcolorbox}[
    colback=blue!6!white,      %
    colframe=black,            %
    coltitle=white,            %
    fonttitle=\bfseries,       %
    boxrule=.7pt,
    width=\linewidth,
    top=1mm,
    bottom=1mm,
    left=2mm,                  %
    right=2mm,                 %
]
\vspace{-4pt}
\textbf{Insight for EcoTab.} In table reasoning, table tokens and text tokens exhibit different uncertainty distributions, and both are informative of the SRM's next-step failure risk.
\vspace{-4pt}
\end{tcolorbox}
\vspace{-9pt}

\section{EcoTab}
\vspace{-5pt}
Motivated by this insight, we propose \textbf{EcoTab}, an efficient table-aware stepwise model routing framework for table reasoning. EcoTab works in three stages. Given the input table $T$, it first builds a lightweight word-level Table Trie to identify table tokens and separate them from text tokens in each reasoning step $s_i$ (Sec.~\ref{sec:3.2}). For each $s_i$, EcoTab then estimates table-token uncertainty $\Phi_{\text{tab}}^{(i)}$ and text-token uncertainty $\Phi_{\text{text}}^{(i)}$ (Sec.~\ref{sec:3.3}). Finally, EcoTab constructs two offline risk mappings, maps the two uncertainties into two failure risks, and combines them into a final routing score $d_{\text{final}}^{(i)}$, which determines whether the next step $s_{i+1}$ should be generated by the SRM or the LRM (Sec.~\ref{sec:3.4}).
\vspace{-5pt}
\subsection{Table Trie Construction}
\vspace{-3pt}
\label{sec:3.2}
To separate table tokens from text tokens, we build a word-level Table Trie from the input table $T$. The Trie stores normalized table content, including column headers and cell values. Before insertion, we apply simple normalization such as lowercasing, removing extra spaces, and standardizing numbers and punctuation. For each reasoning step $s_i = (t_{i,1}, t_{i,2}, \dots, t_{i,k_i})$, we apply the same normalization and scan the step text from left to right with longest-prefix matching over the Trie. If a span matches a column header or cell value, we mark it as table-related tokens. We then map the matched spans back to token positions and obtain a boolean mask $\mathbf{m}^{(i)} \in \{0,1\}^{k_i}$, where $\mathbf{m}^{(i)}_j = 1$ indicates that $t_{i,j}$ is a table token. Based on this mask, we divide the tokens of $s_i$ into a table-token set $\mathcal{V}_{\text{tab}}^{(i)}$ and a text-token set $\mathcal{V}_{\text{text}}^{(i)}$:
\begin{equation}
\mathcal{V}_{\text{tab}}^{(i)} = \{ t_{i,j} \mid \mathbf{m}^{(i)}_j = 1 \},
\mathcal{V}_{\text{text}}^{(i)} = \{ t_{i,j} \mid \mathbf{m}^{(i)}_j = 0 \},
\end{equation}
where $1 \leq j \leq k_i$. This procedure is model-free and introduces little overhead in practice.

\subsection{Step-Level Uncertainty Estimation}
\label{sec:3.3}
Prior studies have shown that reasoning correctness is closely related to model uncertainty \citep{xie2023selfevaluation,wang2024chain}. In particular, the uncertainty of the current step $s_i$ can serve as a useful signal for judging whether the current model has sufficient capability to generate the next step correctly \citep{lee2026confidence}. Specifically, we use Shannon entropy \citep{shannon1948mathematical} as the uncertainty measure for each reasoning step. Each step is generated by the current model, which can be either the LRM or the SRM. To quantify the uncertainty of a reasoning step, we first define token-level uncertainty for each token in the step.


\textbf{Token-Level Uncertainty.}
For the $j$-th token $t_{i,j}$ in reasoning step $s_i$, we define its token-level uncertainty $c_{i,j} \in \mathbb{R}$ as
\begin{equation}
c_{i,j} = - \sum_{v \in \mathcal{V}} p_{i,j}(v)\log p_{i,j}(v),
\vspace{-6pt}
\end{equation}
where $\mathcal{V}$ denotes the model vocabulary, $v \in \mathcal{V}$ is a candidate token set, and $p_{i,j}(v)$ denotes the predicted probability of token set $v$ at position $(i,j)$.

\textbf{Step-Level Uncertainty.}
For each reasoning step $s_i$, we use the Table Trie to partition its tokens into a table-token set $\mathcal{V}_{\text{tab}}^{(i)}$ and a text-token set $\mathcal{V}_{\text{text}}^{(i)}$. We then average the token-level uncertainty separately over the two sets to obtain the table uncertainty $\Phi_{\text{tab}}^{(i)}$ and text uncertainty $\Phi_{\text{text}}^{(i)}$:
\vspace{-4pt}
\begin{equation}
\Phi_{\ast}^{(i)}=
\frac{1}{|\mathcal{V}_{\ast}^{(i)}|}
\sum_{t_{i,j}\in\mathcal{V}_{\ast}^{(i)}} c_{i,j},
\quad \ast \in \{\text{tab}, \text{text}\}.
\vspace{-8pt}
\end{equation}


\vspace{-4pt}

\subsection{Routing Score via Noisy-OR}
\label{sec:3.4}

Given table-token uncertainty $\Phi_{\text{tab}}^{(i)}$ and text-token uncertainty $\Phi_{\text{text}}^{(i)}$, we need to combine them into a single routing score for model selection. However, the two uncertainties follow different distributions. If they are directly averaged or linearly weighted, the resulting score can be poorly calibrated and less reliable for routing. To make them more comparable, we first map them into next-step failure risks in $[0,1]$.
For each uncertainty $\Phi_*^{(i)}$, where $* \in \{\text{tab}, \text{text}\}$, we define
$
d_*^{(i)} \approx \Pr(\text{SRM fails on } s_{i+1} \mid \Phi_*^{(i)}),
$
which measures how likely the SRM is to fail if it is used to generate the next step. To estimate this mapping, we build two offline failure-risk mappings on a held-out validation set. Specifically, we construct step-level supervision by identifying a critical routing boundary for each retained sample, namely the step whose routing score should trigger switching to the LRM for generating the next step. This boundary is determined through suffix replacement on validation trajectories, as detailed in Appendix~\ref{sec:appendix-risk}. The identified boundary step is labeled as positive, earlier retained steps are labeled as negative, and later steps are discarded. This construction aligns the supervision target with the next-step routing objective.
We then fit a sigmoid mapping for each uncertainty:
\begin{equation}
d_*^{(i)} = f_*\!\left(\Phi_*^{(i)}\right)
= \sigma\!\left(a_* \Phi_*^{(i)} + b_*\right),
\vspace{-6pt}
\end{equation}
where $a_*$ and $b_*$ are learned from the validation set. After calibration, the two failure risks are combined using Noisy-OR \citep{pearl2014probabilistic}:
\begin{equation}
d_{\text{final}}^{(i)} = 1 - \left(1 - d_{\text{tab}}^{(i)}\right)\left(1 - d_{\text{text}}^{(i)}\right).
\vspace{-10pt}
\end{equation}
Finally, we compare $d_{\text{final}}^{(i)}$ with a threshold $\tau$ to decide whether the next step $s_{i+1}$ should be generated by the LRM or the SRM.

\subsection{Overall Pipeline}
The overall procedure distinguishes the initial step from all subsequent steps, since no prior routing score is available for the first step. Here, $\theta_M$ and $\theta_m$ denote the LRM and the SRM, respectively.

(1) \textit{Initial step}. For all samples, $\theta_m$ first generates the first reasoning step $s_1$.

(2) \textit{First-step refinement}. Compute the routing score $d_{\text{final}}^{(1)}$ of $s_1$, and regenerate the first step $s_1$ by $\theta_M$ if $d_{\text{final}}^{(1)} > \tau$.

(3) \textit{Iterative}. For each $s_i$, the routing score $d_{\text{final}}^{(i)}$ is used to choose $\theta_M$ or $\theta_m$ for the next step $s_{i+1}$.

(4) \textit{Answer generation}. Repeat step (3) until the final answer is produced or reach the iteration limit.

\vspace{-4pt}

\section{Experiments}
\label{sec:exp}

\vspace{-4pt}
In this section, we present a comprehensive evaluation of EcoTab. We first describe the experimental setup and evaluation metrics (Sec.~\ref{sec5.1}). We then organize our experiments around four key research questions: \textbf{Q1}: Can EcoTab consistently outperform existing state-of-the-art stepwise model routing methods on table reasoning tasks (Sec.~\ref{sec5.2})? \textbf{Q2}: How does each core component of EcoTab contribute to its overall performance (Sec.~\ref{sec5.3})? \textbf{Q3}: Is EcoTab robust to different calibration and threshold settings (Sec.~\ref{sec5.4})? \textbf{Q4}: Can EcoTab effectively capture reasoning difficulty while remaining lightweight in routing overhead (Sec.~\ref{sec5.5})?


\vspace{-2pt}

\subsection{Experimental Setup}
\label{sec5.1}
\begin{table*}[t]
\centering
\resizebox{\textwidth}{!}{
\begin{tabular}{lccccccccccc|>{\columncolor{avggray}}c>{\columncolor{avggray}}c>{\columncolor{avggray}}c}
\toprule
\multirow{2}{*}{\textbf{Method}} & \multirow{2}{*}{\textbf{Extra Tokens}} & \multicolumn{2}{c}{\textbf{WikiTQ}} & \multicolumn{2}{c}{\textbf{TabFact}} & \multicolumn{2}{c}{\textbf{TableBench}} &  \multicolumn{2}{c}{\textbf{HiTab}} & \multicolumn{2}{c}{\textbf{FinQA}}& \multicolumn{3}{c}{\textbf{Average}}  \\
\cmidrule(lr){3-4} \cmidrule(lr){5-6} \cmidrule(lr){7-8} \cmidrule(lr){9-10} \cmidrule(lr){11-12} \cmidrule(lr){13-15} 
 & & Acc $\uparrow$ & FLOPs $\downarrow$ & Acc $\uparrow$ & FLOPs $\downarrow$ & Acc $\uparrow$ & FLOPs $\downarrow$  & Acc $\uparrow$ & FLOPs $\downarrow$&  Acc $\uparrow$ & FLOPs $\downarrow$ & Acc $\uparrow$ & FLOPs $\downarrow$  &A/F $\uparrow$ \\
\midrule
\multicolumn{14}{l}{\textbf{Qwen3-Instruct}}\\
\midrule\midrule
1.7B Only & -- & 67.34 & 3.84 & 82.42 & 2.51 & 42.55 & 6.12 & 59.34 & 4.82 & 47.52 & 7.38 & 59.83 & 4.93 & 15.2\\
14B Only& -- & 84.12 & 21.6 & 90.47 & 14.6 & 54.85 & 42.4  & 80.17 & 20.5 & 67.74 & 48.0 & 75.47 & 29.4 & 3.34\\
\midrule
Random & \xmark      & 75.82 & 18.5 & 86.76 & 10.6 & 46.12 & 36.7 & 73.43 & 17.4 & 58.78 & 40.4 & 68.18 & 24.7 & 3.84\\
RSD& \cmark & 79.73 & 16.2 & 88.24 & 9.32 & 49.21 & 30.4 & 75.88 & 15.2 & 63.87 & 36.4 & 71.39 & 21.5 & 4.55 \\
SpecCoT& \cmark & 78.50 & 15.8 & 87.98 & 8.98 & 49.87 & 29.7 & 76.04 & 15.0 & 64.01 & 36.0 & 71.28 & 21.1 & 4.66 \\
SpecReason& \cmark & 78.76 & 15.7 & 87.74 & 9.21 & 50.02 & 29.9 & 76.41 & 15.0 & \underline{64.21} & 36.0 & 71.43 & 21.1 & 4.62\\
STEER& \xmark & \underline{79.46} & \underline{15.4} & 88.44 & \underline{8.43} & 50.33 & \underline{29.4} & \underline{77.04} & \underline{14.9} & 63.73 & \underline{35.4} & \underline{71.80} & \underline{20.7} & \underline{4.87}\\
GlimpRouter& \cmark & 79.04 & 16.4 & \underline{88.85} & 8.56 & \underline{50.52} & 30.1 & 76.22 & 15.4 & 63.98 & 36.1 & 71.72 & 21.3 & 4.72\\
\rowcolor[HTML]{D8ECE4}
\textbf{EcoTab(ours)}& \xmark & \textbf{80.83} & \textbf{14.0} & \textbf{89.14} & \textbf{8.42} & \textbf{51.77} & \textbf{28.1} & \textbf{78.32} & \textbf{13.9} & \textbf{65.48} & \textbf{34.3} & \textbf{73.11} & \textbf{19.7} & \textbf{5.15}\\
\midrule
\multicolumn{14}{l}{\textbf{DeepSeek-R1-Distill-Qwen}}\\
\midrule\midrule
1.7B Only & -- & 67.34 & 3.84 & 82.42 & 2.51 & 42.55 & 6.12 & 59.34 & 4.82 & 47.52 & 7.38 & 59.83 & 4.93 & 15.22\\
14B Only& -- & 81.71 & 20.8 & 89.21 & 15.3 & 53.65 & 40.5  & 70.08 & 22.9 & 69.40 & 30.9 & 72.81 & 26.0 & 3.28\\
\midrule
Random & \xmark      & 74.78 & 17.9 & 85.77 & 11.1 & 44.03 & 36.4 & 63.06 & 18.1 & 59.67 & 26.7 & 65.46 & 22.0 & 3.77\\
RSD& \cmark & 78.25 & 15.3 & 87.12 & 9.01 & 48.81 & 28.6 & 66.67 & 15.3 & 66.02 & 21.9 & 69.37 & 18.0 & 4.77\\
SpecCoT& \cmark & 78.12 & 15.0 & 87.41 & 8.86 & 48.74 & 28.3 & 66.94 & 15.0 & 65.89 & 22.6 & 69.42 & 17.9 & 4.83\\
SpecReason& \cmark & 78.01 & 15.0 & 87.57 & 8.94 & 48.89 & 28.4 & 66.58 & 15.1 & 66.07 & 22.2 & 69.42 & 17.9 & 4.82\\
STEER& \xmark & \underline{78.32} & \underline{14.4} & 87.64 & 8.83 & 49.22 & 29.1 & \underline{67.01} & 15.3 & \underline{66.15} & \textbf{21.3} & \underline{69.67} & 17.8 & \underline{4.91}\\
GlimpRouter& \cmark & 78.04 & 14.8 & \underline{87.81} & \textbf{8.78} & \underline{49.54} & \underline{28.1} & 66.71 & \underline{14.8} & 65.81 & 22.8 & 69.58 & \underline{17.8} & 4.89\\
\rowcolor[HTML]{D8ECE4}
\textbf{EcoTab(ours)}& \xmark & \textbf{79.56} & \textbf{13.1} & \textbf{88.04} & \underline{8.81} & \textbf{50.41} & \textbf{27.4} & \textbf{67.97} & \textbf{13.9} & \textbf{67.63} & \underline{21.4} & \textbf{70.72} & \textbf{16.9} & \textbf{5.19} \\

\bottomrule
\end{tabular}%
}
\caption{
Main Results. {\footnotesize\underline{Acc $\uparrow$}} is measured at 60\% of LRM-only FLOPs, and {\footnotesize\underline{FLOPs $\downarrow$}} denotes the computation required to reach 98\% of LRM-only accuracy. The best result is \textbf{bold}, and the second-best result is \underline{underlined}. Extra Tokens denotes whether the method requires additional token generation during the reasoning process.
}
\label{tab:main-exp2}
\vspace{-14pt}
\end{table*}


\paragraph{Models and Configurations.}
We use Qwen3-1.7B \citep{yang2025qwen3} as the SRM, and evaluate two LRMs: Qwen3-14B \citep{yang2025qwen3} and DeepSeek-R1-Distill-Qwen-14B \citep{guo2025deepseek}. This setup allows us to study both same-family and cross-family collaboration.

\paragraph{Benchmarks.} We evaluate EcoTab on five table reasoning benchmarks. 
TableBench \citep{wu2025tablebench} contains 886 complex questions spanning numerical reasoning, fact checking, and data analysis. 
WikiTQ \citep{pasupat-liang-2015-compositional} focuses on question answering over Wikipedia tables, and TabFact \citep{Chen2020TabFact:} evaluates table-based fact verification. 
To further test robustness, we include HiTab \citep{cheng-etal-2022-hitab}, which features hierarchical nested tables, and FinQA \citep{chen-etal-2021-finqa}, a text+table reasoning dataset requiring joint understanding of financial reports and tabular data. 
Additional details are provided in Appendix \ref{sec:a2}.

\vspace{-4pt}

\paragraph{Baselines.} We compare EcoTab against standalone models and state-of-the-art stepwise model routing baselines, including SRM/LRM Only, Random, RSD \citep{liao2025rewardguided}, SpecCoT \citep{shi-etal-2025-speccot}, SpecReason \citep{pan2025specreason}, STEER \citep{lee2026confidence} and GlimpRouter \citep{zeng2026glimprouter}.

\vspace{-4pt}

\paragraph{Evaluation Metric.}
Following prior work \citep{liao2025rewardguided,sardana2024beyond}, we use accuracy as the performance metric and estimate inference cost using the standard Transformer FLOPs approximation of \(2N\) per generated token for a model with \(N\) parameters. To enable clearer comparison across baselines, we conduct a grid search over threshold values with a step size of 0.05 for all baseline methods and EcoTab, and derive the corresponding accuracy--FLOPs trade-off curves. We then report the accuracy achieved at \(60\%\) of the LRM-only FLOPs, as well as the FLOPs required to reach \(98\%\) of the LRM-only accuracy. In addition, we report Accuracy-per-FLOPs (A/F), adapted from \citep{ma2025cot}, to better characterize the trade-off between performance and computational cost.



\begin{table}[t]
\centering
\resizebox{0.48\textwidth}{!}{
\begin{tabular}{l|cc|cc}
\toprule
\textbf{Method} & \textbf{Acc $\uparrow$} & $\triangledown$ & \textbf{FLOPs $\downarrow$} & $\triangledown$ \\
\midrule 
\textbf{EcoTab (Qwen3-Instruct)} & \textbf{51.8} & -- & \textbf{28.1} & -- \\
\quad w/ average token & 49.9 & \textcolor[HTML]{CC0000}{(-1.9)} & 30.3 & \textcolor[HTML]{CC0000}{(+2.2)} \\
\quad w/ only table token & 50.5 & \textcolor[HTML]{CC0000}{(-1.3)} & 29.4 & \textcolor[HTML]{CC0000}{(+1.3)} \\
\quad w/ only text token & 50.2 & \textcolor[HTML]{CC0000}{(-1.6)} & 29.7 & \textcolor[HTML]{CC0000}{(+1.6)} \\
\quad w/ linear weighting & 50.0 & \textcolor[HTML]{CC0000}{(-1.8)} & 29.8 & \textcolor[HTML]{CC0000}{(+1.7)} \\
\midrule
\textbf{EcoTab (DeepSeek-R1)} & \textbf{50.4} & -- & \textbf{27.4} & -- \\
\quad w/ average token & 48.9 & \textcolor[HTML]{CC0000}{(-1.5)} & 30.0 & \textcolor[HTML]{CC0000}{(+2.6)} \\
\quad w/ only table token & 49.6 & \textcolor[HTML]{CC0000}{(-0.8)} & 28.3 & \textcolor[HTML]{CC0000}{(+0.9)} \\
\quad w/ only text token & 49.4 & \textcolor[HTML]{CC0000}{(-1.0)} & 28.6 & \textcolor[HTML]{CC0000}{(+1.2)} \\
\quad w/ linear weighting & 49.1 & \textcolor[HTML]{CC0000}{(-1.3)} & 29.0 & \textcolor[HTML]{CC0000}{(+1.6)} \\
\bottomrule
\end{tabular}
}
\caption{Ablation results of EcoTab on TableBench. We report accuracy (Acc) and the average FLOPs per query.}
\label{tab:ablation1}
\vspace{-16pt}
\end{table}

\subsection{Main Results}
\label{sec5.2}
Table~\ref{tab:main-exp2} presents the main results of EcoTab on five table reasoning benchmarks under both same-family and cross-family collaboration settings. Overall, EcoTab consistently outperforms all baselines and achieves the best accuracy--efficiency trade-off in both settings. In particular, it obtains the best overall average performance in terms of accuracy, FLOPs, and A/F, showing that EcoTab can improve reasoning quality while reducing inference cost. A closer look shows that EcoTab remains consistently effective across different datasets and model combinations. Under both Qwen3-14B and DeepSeek-R1-Distill-Qwen-14B as the LRM, EcoTab achieves the highest accuracy on all five benchmarks, while also maintaining the lowest overall inference cost. This result verifies that the advantage of EcoTab is not limited to a specific model family, but generalizes to both same-family and cross-family collaboration. Another notable advantage is that EcoTab does not require extra token generation during routing, yet it still surpasses strong baselines that rely on verification overhead. This suggests that explicitly separating table-token and text-token is more effective for table reasoning than directly applying free-form text routing methods. Overall, the results confirm that EcoTab is a more suitable step-level routing framework for table reasoning, delivering a stronger balance between effectiveness and efficiency.

\subsection{Ablation Study}
\label{sec5.3}
Table~\ref{tab:ablation1} reports the ablation results of EcoTab on TableBench. Overall, the full EcoTab consistently performs best under both Qwen3-Instruct and DeepSeek-R1 settings, achieving the highest accuracy with the lowest FLOPs. Removing the separation between table tokens and text tokens (\emph{w/ average token}) causes the largest performance drop and a clear increase in inference cost, showing that collapsing all tokens into a single score weakens routing decisions. Using only table tokens (\emph{w/ only table token}) or only text tokens (\emph{w/ only text token}) also degrades performance, indicating that both types of token are necessary and complementary for effective routing. In addition, replacing Noisy-OR with simple linear weighting (\emph{w/ linear weighting}) consistently hurts both accuracy and efficiency, suggesting that EcoTab benefits not only from separating the two token types, but also from using a more suitable fusion strategy. Overall, these results indicate that all components of EcoTab are essential to its effectiveness.

\begin{figure}[t]
    \centering
    \includegraphics[width=1\linewidth]{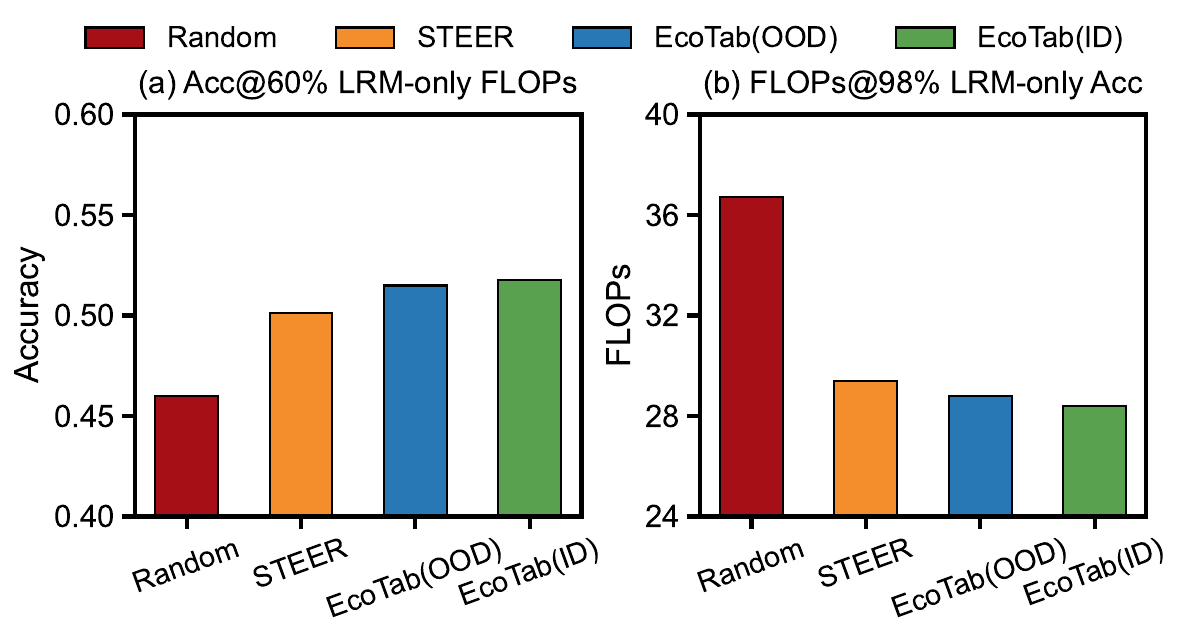}
    \caption{Failure-risk mapping transferability of EcoTab. ID denotes in-domain evaluation, and OOD denotes out-of-domain transfer.}
    \label{fig:exp1}
    \vspace{-14pt}
\end{figure}

\subsection{Robustness Analysis}
\label{sec5.4}
\paragraph{Failure-risk mapping transferability.} We study whether the fitted score-to-risk mapping in Sec.~\ref{sec:3.4} can transfer across domains. Specifically, we compare an in-domain setting (ID), where the mapping is fitted on the target dataset, with an out-of-domain setting (OOD), where the mapping fitted on the WikiTQ validation split is directly applied to TableBench. Figure~\ref{fig:exp1} shows that the OOD variant still outperforms Random and STEER on both metrics. It achieves higher accuracy at 60\% of LRM only FLOPs and requires fewer FLOPs to reach 98\% of LRM only accuracy. Although the ID setting performs slightly better, the gap is small, which suggests that the learned risk mapping generalizes well across table reasoning domains.

\vspace{-4pt}

\begin{figure}[t]
    \centering
    \includegraphics[width=0.95\linewidth]{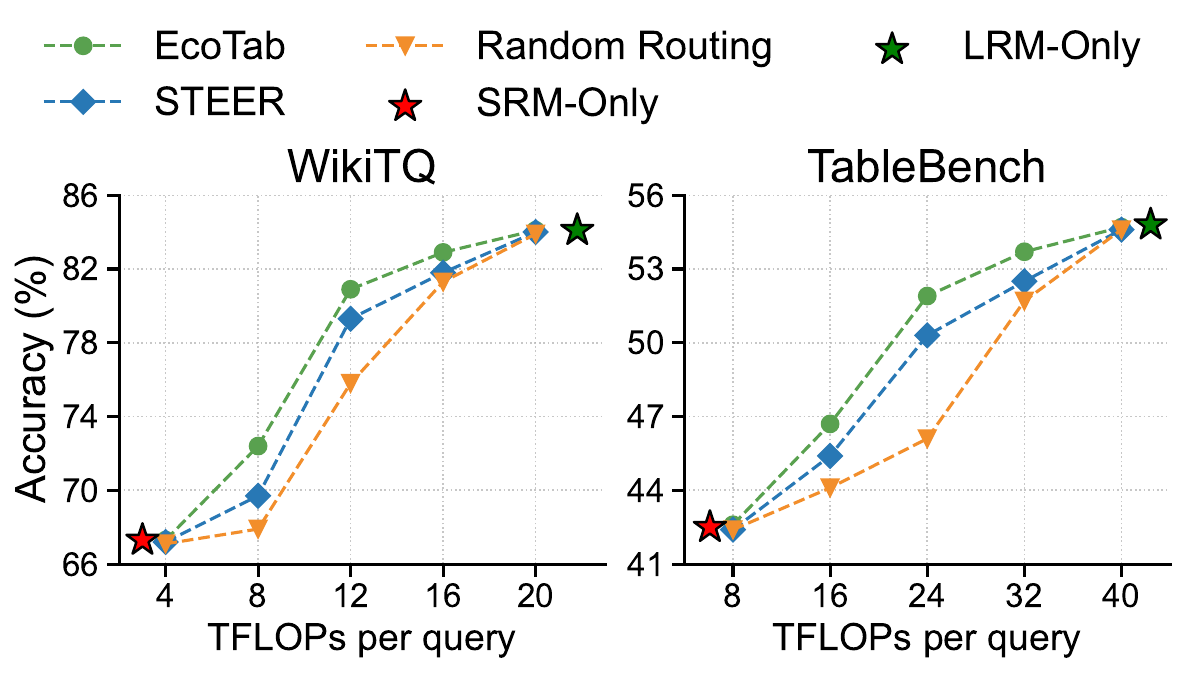}
    \caption{Threshold robustness of EcoTab on WikiTQ and TableBench. As the threshold $\tau$ increases, the total FLOPs decrease.}
    \label{fig:exp2}
    \vspace{-14pt}
\end{figure}

\paragraph{Threshold robustness.} We further vary the routing threshold and plot the accuracy--FLOPs curves on WikiTQ and TableBench. A higher threshold makes the router more conservative in calling the LRM, so fewer steps are assigned to the LRM and the total FLOPs become lower. As shown in Figure~\ref{fig:exp2}, EcoTab consistently stays above STEER and Random on both datasets. It achieves higher accuracy at similar cost, or lower cost at similar accuracy. Overall, the gain comes from more reliable model routing rather than a specific threshold.

\vspace{-4pt}

\subsection{Further Analysis}
\vspace{-2pt}
\label{sec5.5}
\paragraph{Difficulty awareness.}

\begin{figure}[t]
    \centering
    \includegraphics[width=0.99\linewidth]{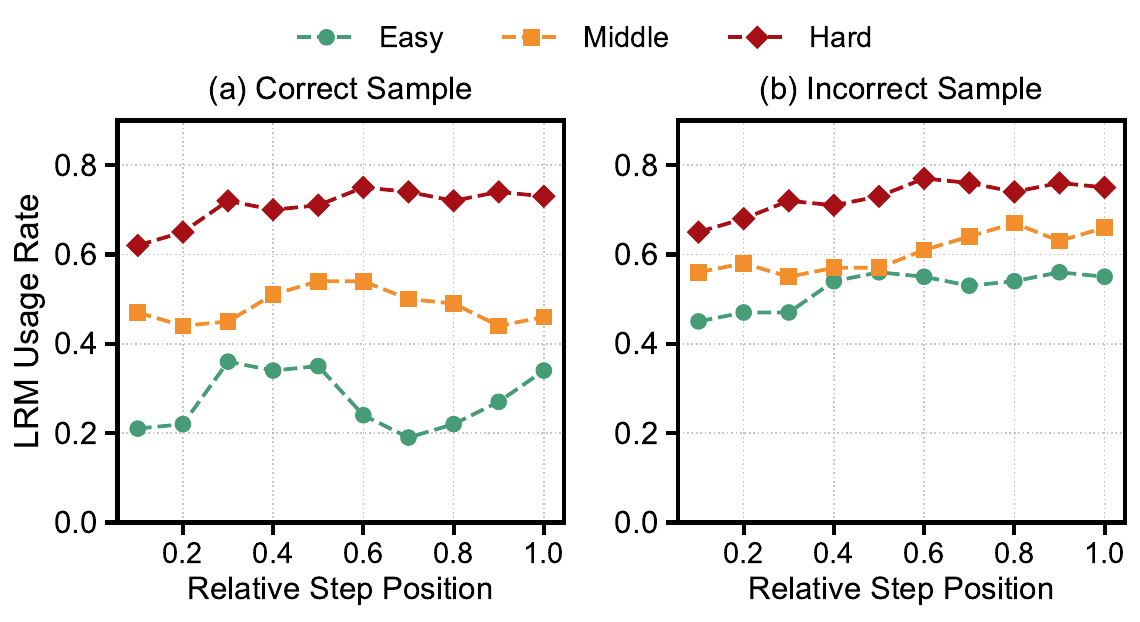}
    \caption{LRM usage rate across GPT-5.4-High difficulty levels, comparing correct and incorrect samples.}
    \label{fig:exp3}
    \vspace{-10pt}
\end{figure}

We further examine whether EcoTab is sensitive to sample difficulty. Following prior work~\citep{ye2025tableqa}, we estimate TableBench difficulty using GPT-5.4-High. For each question, we sample 100 independent answers and use the number of correct responses as a proxy for difficulty. Based on this score, we group samples into three subsets: \textit{hard} (0--9 correct), \textit{medium} (10--59 correct), and \textit{easy} (60--100 correct). We then compute the LRM usage rate of EcoTab at different relative step positions within each group.
Figure~\ref{fig:exp3} shows that harder samples consistently trigger more LRM calls than easier ones. This pattern holds for both correct and incorrect samples, and is especially clear on incorrect samples. These results suggest that EcoTab can capture sample-level difficulty and allocate more LRM computation to harder cases.

\begin{figure}[t]
    \centering
    \includegraphics[width=1\linewidth]{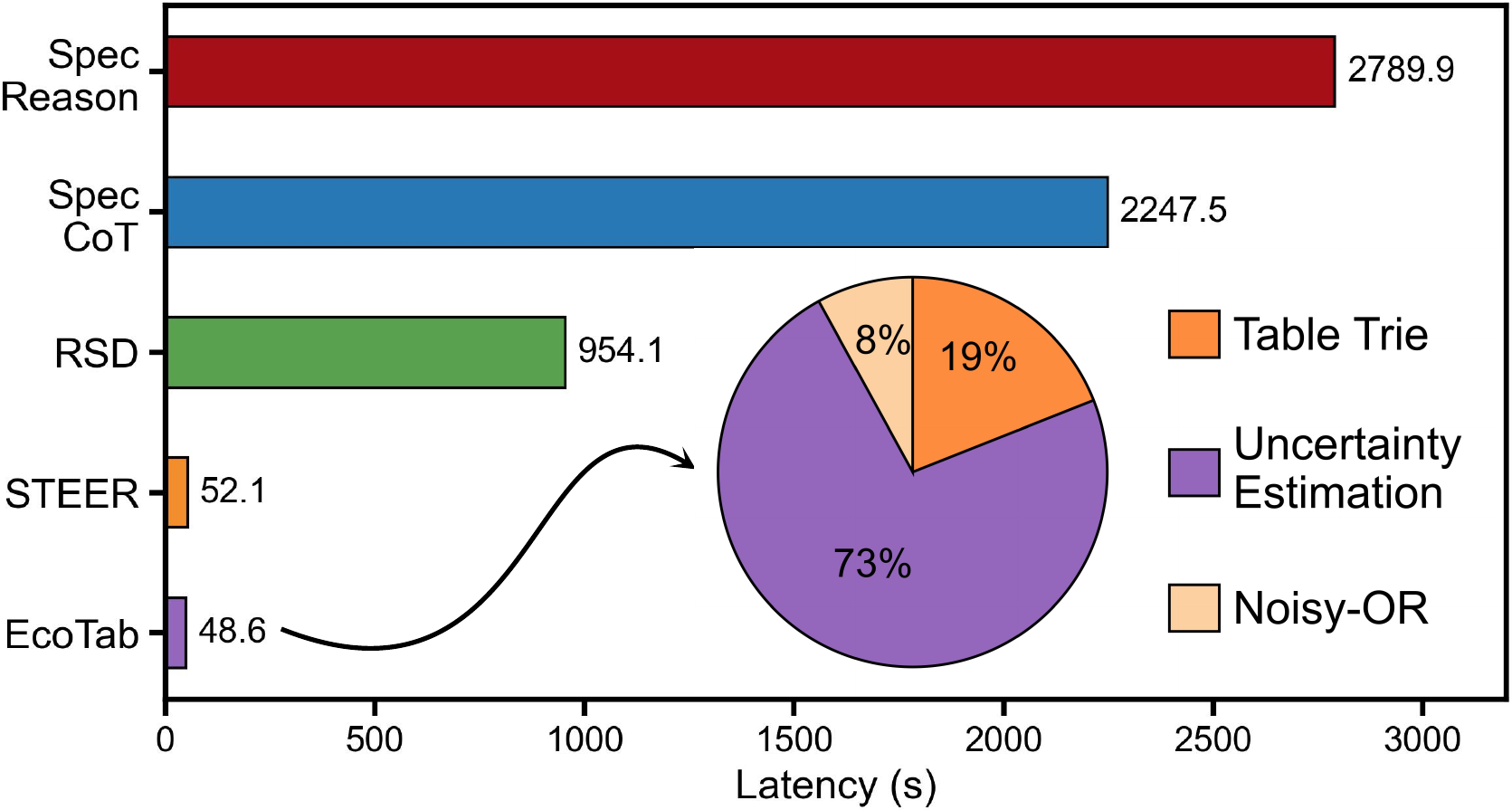}
    \caption{Overall routing latency on TableBench, together with the latency breakdown of EcoTab.}
    \label{fig:exp4}
    \vspace{-10pt}
\end{figure}

\paragraph{Routing overhead.} We then compare the routing latency of different methods on the full TableBench benchmark, excluding the reasoning time and measuring only model routing overhead. Figure~\ref{fig:exp4} shows that EcoTab has an overall latency of 48.6 seconds, which is comparable to STEER at 52.1 seconds, while remaining far below RSD, SpecCoT, and SpecReason. This result shows that EcoTab introduces little extra overhead and remains a lightweight routing method in practice.

\section{Related Work}
\paragraph{Table Reasoning.} 
Reasoning over tables poses unique challenges for LLMs, as it requires both natural language understanding and structured reasoning over rows, columns, and cell values \citep{jin2022survey,zhang2025survey}. Recent studies \citep{chen2020hybridqa,deng2022turl,iida2021tabbie} have explored table reasoning across a variety of downstream tasks, including table qa \citep{chen2020hybridqa} and table fact verification \citep{parikh2020totto}. Early methods, such as TAPAS \citep{herzig2020tapas} and TaBERT \citep{yin2020tabert}, mainly model tables through Transformer-based encoders. With the rise of LLMs, later approaches began to improve table reasoning through prompt engineering \citep{sui2024tap4llm,wang2024chainoftable} or supervised fine-tuning \citep{su2024tablegpt2}. More recent works, such as the Table-R1 series \citep{yang2025table,wu2025table,jin2025table}, further enhance reasoning performance by using reinforcement learning to optimize reasoning trajectories. However, as table size grows and reasoning models become larger and more verbose, inference latency becomes increasingly prohibitive. EcoTab addresses this issue by introducing a table-aware step-level routing framework to better balance efficiency and performance.


\paragraph{Efficient Reasoning.}
Scaling test-time compute in LRMs can substantially improve reasoning performance, but also incurs prohibitive latency \citep{wang2025vrag}. To mitigate this cost, recent work has explored dynamically offloading part of the reasoning process to smaller models, mainly at three levels: query-level routing \citep{chen2023frugalgpt,wang2025mixllm,zhao2025optimizing}, step-level routing \citep{shi-etal-2025-speccot,pan2025specreason,lee2026confidence}, and token-level speculation \citep{leviathan2023fast,chen2023accelerating}. Among them, step-level routing is particularly well aligned with the multi-step reasoning nature of LRMs. Existing methods range from training-based reward guidance, such as RSD \citep{liao2025rewardguided}, to training-free paradigms based on multi-path selection, such as SpecCoT \citep{shi-etal-2025-speccot}, post-hoc verification, such as SpecReason \citep{pan2025specreason}, and methods that route models based on uncertainty signals, such as entropy \citep{zeng2026glimprouter} or confidence \citep{lee2026confidence}. However, these methods are mainly designed for free-form text reasoning and do not account for the structured nature of table reasoning. 


\vspace{-4pt}

\section{Conclusion}
\vspace{-4pt}
In this paper, we study efficient stepwise model routing for table reasoning with LRMs. We show that existing step-level routing methods designed for free-form text reasoning are less effective in tabular settings due to the structured nature of tables. To address this issue, we propose EcoTab, an adaptive table-aware step-level routing framework that explicitly distinguishes table tokens and text tokens when estimating step difficulty. The resulting routing score is computed through a probabilistic fusion mechanism to guide model selection during reasoning. 
Experiments on multiple table reasoning benchmarks demonstrate that EcoTab consistently achieves a better trade-off between reasoning accuracy and computational cost compared with existing stepwise model routing methods. These results highlight the importance of table-aware routing for efficient reasoning over structured data.

\section*{Limitations}
Like existing routing methods, EcoTab cannot explicitly control the output length of the reasoning process. In practice, the routing decision only determines which model generates the next step, but does not decide when the reasoning should stop. As a result, unnecessary long reasoning traces may still appear even when the routing is accurate. A promising direction for future work is to design a table aware early stopping mechanism that can terminate reasoning once the required table evidence and logical deductions are sufficient.

\vspace{-4pt}

\section*{Ethics Statement}

Our work aims to improve the efficiency and reliability of multi-step table reasoning through stepwise model routing. However, like any system built on LLMs, it may still produce incorrect intermediate reasoning steps or factually incorrect final answers. We therefore encourage users to exercise caution and verify critical outputs when deploying such systems in real-world scenarios. Furthermore, our research builds upon open-source models and frameworks, including Qwen3, DeepSeek-R1-Distill, PyTorch, and Hugging Face. We strictly follow their respective licenses and usage policies, and acknowledge their important contributions to the research community.


\section*{Acknowledgement}

This work is supported by the Fundamental Research Funds for the Central Universities under Grant No. WK2102026004 and Anhui Provincial Science and Technology Innovation Key Research Program (202523j08050016).

\vspace{-8pt}

\bibliography{custom}

\appendix

\section{Additional Experimental Setups}
\label{sec:appendix1}

\subsection{Model Configurations}
\label{sec:a1}
In our experiments, we use Qwen3-1.7B \citep{yang2025qwen3} as the SRM. For the LRM, we consider two representative settings: Qwen3-14B \citep{yang2025qwen3} for same-family collaboration and DeepSeek-R1-Distill-Qwen-14B \citep{guo2025deepseek} for cross-family collaboration. This design allows us to evaluate whether EcoTab remains effective under both homogeneous and heterogeneous model pairs.
For EcoTab and all compared baselines, we use the same decoding configuration for fair comparison. Specifically, we set the temperature to 0.7, the maximum generation length to 16{,}384 tokens, and top-$p$ sampling with $p=0.95$.

\subsection{Dataset Details}
\label{sec:a2}
\paragraph{TabFact \citep{Chen2020TabFact:}.}
TabFact is a large-scale benchmark for table-based fact verification rather than standard table question answering. It consists of approximately 16K Wikipedia tables paired with 118K human-annotated natural language statements, where each statement is labeled as either \textit{ENTAILED} or \textit{REFUTED} with respect to the corresponding table. The dataset is challenging because it requires not only semantic understanding of natural language statements, but also symbolic reasoning over semi-structured tables, such as comparison, counting, and aggregation.

\vspace{-4pt}

\paragraph{WikiTableQuestions (WikiTQ) \citep{pasupat-liang-2015-compositional}.}
WikiTQ is a benchmark for answering complex natural language questions over semi-structured HTML tables. It contains 22,033 question-answer pairs over 2,108 Wikipedia tables. A key characteristic of WikiTQ is that the training and test tables are disjoint, which requires models to generalize to unseen table schemas. The questions often involve compositional reasoning, including comparison, superlatives, aggregation, and arithmetic operations. The tables are semi-structured and non-normalized, and many cells contain multi-part values that must be interpreted appropriately during reasoning.

\vspace{-4pt}

\paragraph{TableBench \citep{wu2025tablebench}.}
TableBench is a comprehensive and challenging benchmark designed to evaluate complex table question answering in more realistic scenarios. It covers 18 fine-grained subcategories under four major categories, namely fact checking, numerical reasoning, data analysis, and visualization, with a total of 886 benchmark instances. The benchmark is built from 3,681 unique tables spanning diverse domains, with an average of 16.71 rows and 6.68 columns per table. In addition, 65.74\% of table cells are numerical, and each instance requires 6.26 reasoning steps on average, making TableBench substantially more difficult than earlier TableQA benchmarks.

\vspace{-4pt}

\paragraph{HiTab \citep{cheng-etal-2022-hitab}.}
HiTab is a benchmark for question answering and natural language generation over hierarchical tables. Unlike prior datasets that mainly focus on flat tables, HiTab emphasizes hierarchical indexing and implicit semantic and numerical relations induced by table structure. It is a cross-domain dataset constructed from statistical reports and Wikipedia pages, and nearly all tables exhibit hierarchical organization. The dataset contains 10,686 QA pairs and descriptive sentences over 3,597 tables, together with fine-grained annotations of entity and quantity alignment, which make it suitable for studying complex reasoning over hierarchical tabular data.

\vspace{-4pt}

\paragraph{FinQA \citep{chen-etal-2021-finqa}.}
FinQA is a financial-domain dataset for complex numerical reasoning over heterogeneous evidence. It contains 8,281 question-answer pairs annotated by finance professionals, along with gold reasoning programs for explainable evaluation. The dataset is constructed from earnings reports of S\&P 500 companies and requires models to integrate information from both tables and accompanying unstructured text. Compared with general-domain TableQA benchmarks, FinQA places greater emphasis on multi-step numerical reasoning and domain-specific financial knowledge.

\vspace{-4pt}

\paragraph{Validation set in Sec.~\ref{sec:3.4}.}
The held-out validation set in Sec.~\ref{sec:3.4} is used only for fitting the offline risk mappings and is strictly separated from the final evaluation set. For datasets with an official training split, we construct the validation set solely from the training data. For datasets without a predefined training split, we randomly sample 10\% of the original test set as a pseudo-validation split, and use the remaining 90\% for final testing. In all cases, the validation data are used only for calibration and are never included in the final reported test results.
\vspace{-3pt}

\subsection{Implementation Details}
\label{sec:a3}
\vspace{-3pt}
For the accuracy--FLOPs evaluations in Sec.~\ref{sec2} and Sec.~\ref{sec:exp}, we follow a unified threshold-sweeping protocol for EcoTab and all compared baselines. Specifically, for each method, we perform a grid search over the routing threshold $\tau$ with a step size of 0.05, and compute the corresponding accuracy and average FLOPs under different values of $\tau$. This process produces the full accuracy--FLOPs trade-off curve for each method. For the main results in Table~\ref{tab:main-exp2}, we report two representative operating points from the trade-off curve. First, we use the accuracy achieved at 60\% of the LRM-only FLOPs as the reported \textbf{Acc}. Second, we use the FLOPs required to reach 98\% of the LRM-only accuracy as the reported \textbf{FLOPs}. These two metrics respectively reflect the model quality under a fixed computation budget and the computation required to approach near-LRM performance.
In addition, we report Accuracy-per-FLOPs (A/F) as an overall indicator of the effectiveness--efficiency trade-off \citep{ma2025cot}. Following our main experimental setup, FLOPs are estimated using the standard Transformer approximation of $2N$ per generated token for a model with $N$ parameters. All reported results are averaged over three independent runs.

\subsection{Four Error Types for Failed Steps}

Following TATTOO~\citep{zou2026tattoo}, we manually inspect the failed trajectories and categorize each failed step into one of four error types.

\textbf{Table Retrieval Step.}
This type includes row or column mis-selection, unit mismatch, and partial aggregation errors. These errors account for 47.7\% of all failed steps, indicating that a substantial portion of failures arise from difficulty in correctly locating and extracting the relevant table region.

\textbf{Table Operation Step.}
This type covers miscalculation, grouping mistakes, double counting, and misinterpretation of table semantics. It represents 34.3\% of all failed steps, suggesting that even after the relevant contents are retrieved, reasoning over structured tabular information remains challenging.

\textbf{Inner-Thinking Step.}
This type refers to logical mistakes or self-contradictory reasoning that are not directly caused by table grounding. Such errors account for 12.0\% of all failed steps, indicating that LRMs are relatively more reliable on pure logical chains than on table-centric operations.

\textbf{Others.}
This category includes failures caused by context omission, incomplete responses, or improper output formatting.

To provide a more concrete understanding of these error types, Table~\ref{tab:error_example} presents representative failure cases from three major categories: \textit{Table Retrieval}, \textit{Table Operation}, and \textit{Inner-Thinking}. For each case, we show the first erroneous reasoning step in the trajectory and briefly explain how this mistake propagates to the final incorrect answer. These examples illustrate that failures in table reasoning may arise from different stages of the reasoning process, including incorrect table grounding, faulty operations over retrieved contents, and purely logical mistakes.


\vspace{-6pt}

\section{Table Trie Construction Details}
\label{sec:appendix-trie}
\vspace{-4pt}
To separate table tokens from text tokens in each reasoning step, EcoTab builds a word-level Table Trie from the input table. The trie stores normalized textual entries extracted from the table, including column headers and cell values. During inference, each reasoning step is normalized in the same way and scanned from left to right with longest-prefix matching. The matched spans are then mapped back to token positions to obtain the table-token mask used in Eq.~(4). This implementation follows the procedure described in Sec.~\ref{sec:3.2}.
\vspace{-2pt}
\subsection{Normalization Rules}
\vspace{-2pt}
We apply lightweight normalization to both table contents and reasoning steps before trie construction and matching. The goal is to improve robustness to surface-form variation while keeping the procedure simple and efficient.


\textbf{Lowercasing and whitespace cleanup.}
All text is converted to lowercase. Consecutive spaces, tabs, and line breaks are collapsed into a single space, and leading or trailing whitespace is removed.


\textbf{Punctuation normalization.}
Common punctuation variants are standardized into a unified form. For example, different dashes and quotation marks are mapped to their canonical ASCII forms when possible. Surrounding punctuation that does not affect semantic identity is ignored during matching.


\textbf{Number normalization.}
We normalize common number formats to reduce mismatches caused by formatting differences. For example, ``1,200'' and ``1200'' are treated as the same value. Decimal numbers and percentages are preserved in normalized form when they carry semantic meaning.


\textbf{Cell and header insertion.}
We insert both column headers and cell values into the trie. Each entry is split at the word level after normalization. Multi-word entries such as ``new york'' or ``gross domestic product'' are inserted as complete paths rather than as isolated words.


\textbf{Token-span consistency.}
Normalization is applied only for matching. After a span is matched in normalized text, we map it back to the original token positions in the reasoning step, so the final boolean mask is still defined over the original generated tokens.

\begin{table*}[t]
    \centering
    \small
    \resizebox{\linewidth}{!}{%
    \begin{tabular}{l|>{\small}p{0.6\linewidth}|>{\small}p{0.5\linewidth}}
    \specialrule{1.2pt}{0pt}{0pt} 
    \textbf{Category} & \textbf{Wrong Model Answer} & \textbf{Error Description}\\
    \midrule
    \multirow{5}{*}{Table Retrieval} & 
    <think> Step 1: I need to figure out the average prominence of mountain peaks in the Democratic Republic of the Congo that have an elevation of at least 3000 meters. \newline \highlightred{Step 2: First, I'll review the provided table. My main focus is on the `country' and `elevation (m)' columns. \textcolor{red}{\textit{(Wrong Step)}}} \dots </think> \newline Final Answer is \texttt{\textbackslash boxed\{2306.65\}}
    & The model initially retrieves the wrong part of the table (columns \textit{`Country'} and \textit{`Elevation (m)'}), which leads to errors in all subsequent reasoning steps.
    \\
    \midrule
    \multirow{5}{*}{Table Operation} & 
    <think> Step 1: Let me start by looking at the table provided \dots The retrieved columns are \dots \newline \highlightred{Step 4: I'll go through each row and note down the episodes: Episode 1: Rating - 5.1, Episode 2: Rating - 5.2 \dots \textcolor{red}{\textit{(Wrong Step)}}} Step 5: Adding the rating together is 5.1 + 5.2 + \dots </think> \newline Final Answer is: \texttt{\textbackslash boxed\{11.52\}} 
    & The model identifies the correct column for the answer, but fails to extract all elements, omitting the last cell. This omission leads to incorrect reasoning in all subsequent steps.
    \\
    
    \midrule
    \multirow{5}{*}{Inner-thinking} & 
    <think> Step 1: Based on the table contents, I need to count podiums per season. 2008: $(1+1+1)/3 = 1$, \; 2009: $6$, \; 2010: $0$, \; 2011: $(2+0)/2 = 1$ \dots \newline \highlightred{Step 4: Then I will average across seasons. Final average is $(1+6+0+1)/4 = 2$  \textcolor{red}{\textit{(Wrong Step)}}} </think>  \newline Final Answer is: \texttt{\textbackslash boxed\{2\}}
    & The model incorrectly does the calculation by averaging the season-level means, giving each season equal weight, instead of averaging across all team-seasons.
    \\
    
    \specialrule{1.2pt}{0pt}{0pt} 
    \end{tabular}
    }
    \caption{Representative error cases for major failed-step categories.}
    \label{tab:error_example}
    \vspace{-10pt}
\end{table*}

\subsection{Trie Matching Algorithm}

Given a reasoning step $s_i = (t_{i,1}, t_{i,2}, \dots, t_{i,k_i})$, we first normalize the step text using the same rules as above. We then scan the step from left to right and perform longest-prefix matching over the trie.
At each position, we attempt to extend the current span word by word along the trie. If multiple matches are possible, we keep the longest valid match. For example, if both ``new'' and ``new york'' exist in the trie, the matcher prefers ``new york'' whenever the longer span is observed in the step. Once a match is confirmed, all tokens covered by that span are marked as table-related, and the scan continues from the end of the matched span. If no match is found, the scan advances by one token.
Formally, this process returns a boolean mask $\mathbf{m}^{(i)} \in \{0,1\}^{k_i}$, where $\mathbf{m}^{(i)}_j = 1$ indicates that token $t_{i,j}$ belongs to a matched table-related span. Based on this mask, the step tokens are partitioned into the table-token set $V_{\text{tab}}^{(i)}$ and the text-token set $V_{\text{text}}^{(i)}$, which are then used to compute $\Phi_{\text{tab}}^{(i)}$ and $\Phi_{\text{text}}^{(i)}$ in Eq.~(4). Although this matching is surface-form based, lightweight normalization already covers manycommon variations in numbers and punctuation, and we find it sufficient in practice.

\section{Failure-risk Mapping Implementation}
\label{sec:appendix-risk}

\subsection{Construction of Fitting Data}

EcoTab performs next-step model routing: the routing score computed from the current step $s_i$ is used to decide whether the next step $s_{i+1}$ should be generated by the SRM or the LRM. To align the supervision target with this objective, we construct step-level labels by identifying a critical routing boundary through counterfactual suffix replacement.

For each dataset, we first build a held-out validation split following Appendix~\ref{sec:a2}, and run both LRM-only and SRM-only inference to obtain full reasoning trajectories. We retain only samples that are correct under LRM-only but incorrect under SRM-only, since these cases directly indicate that the stronger model is needed for part of the reasoning process. Each trajectory is segmented into steps using the delimiter ``\textbackslash n\textbackslash n''. For every step $s_i$ in the retained LRM trajectory, we compute the table-token uncertainty $\Phi_{\text{tab}}^{(i)}$ and the text-token uncertainty $\Phi_{\text{text}}^{(i)}$.
For a retained trajectory with steps $s_1,\dots,s_T$, we progressively replace its suffix with SRM generations, starting from the last step. For a suffix length $m \in \{1,\dots,M\}$ with $M=\min(T-1,8)$, we keep the prefix $s_1,\dots,s_{T-m}$ fixed and let the SRM regenerate the remaining suffix. For each $m$, we repeat generation $k=5$ times and evaluate the final answer. A suffix is regarded as causing a stable outcome flip if at least $4$ out of the $5$ runs become incorrect, corresponding to a flip ratio of at least $\gamma=0.8$.

We then define $m^\star$ as the smallest suffix length that causes a stable outcome flip, and stop the search immediately once it is found. The corresponding critical routing boundary is $b=T-m^\star$. The step $s_b$ is labeled as a positive sample because its routing score should have triggered the switch to the LRM for generating the next step. All earlier steps $s_1,\dots,s_{b-1}$ are labeled as negative, and all later steps are discarded. If no stable outcome flip is found within the scanned range, the sample is excluded from the fitting set.
This construction provides step-level supervision that is better aligned with the next-step routing objective than assigning labels directly from the final trajectory outcome. Since the table-token and text-token mappings are fitted on the same retained step set, they share the same binary labels and the same total number of samples. Table~\ref{tab:risk-data-size} reports the total number of retained step-level samples under this construction.

\begin{table}[t]
\centering
\small
\begin{tabular}{lc}
\toprule
Dataset & Retained samples $S$ \\
\midrule
WikiTQ & 2281 \\
TabFact & 2374 \\
TableBench & 432 \\
HiTab & 1742 \\
FinQA & 1165 \\
\bottomrule
\end{tabular}
\caption{Number of retained step-level samples used for fitting the offline risk mappings under the suffix-replacement construction.}
\label{tab:risk-data-size}
\vspace{-14pt}
\end{table}

\vspace{-4pt}

\subsection{Fitting Failure-risk Mappings}
\vspace{-4pt}
For each dataset, we fit two independent sigmoid risk mappings, one for table-token uncertainty and the other for text-token uncertainty, using the retained step-level samples constructed above. For each signal $\Phi_*^{(i)}$, where $* \in \{\text{tab}, \text{text}\}$, the risk score is defined as
\vspace{-3pt}
\begin{equation}
d_*^{(i)} = f_*\!\left(\Phi_*^{(i)}\right) = \sigma\!\left(a_* \Phi_*^{(i)} + b_*\right),
\end{equation}

where $\sigma(\cdot)$ is the sigmoid function, and $a_*$ and $b_*$ are learned from the retained validation samples of that dataset. Let
\vspace{-3pt}
\[
p = \sigma(a_* \Phi + b_*).
\]
The fitting objective is the standard binary cross-entropy:
\begin{equation}
\mathcal{L}_* = -\sum_{(\Phi,y)\in\mathcal{D}_*} \left[ y\log p + (1-y)\log(1-p) \right].
\end{equation}

During inference, EcoTab maps $\Phi_{\text{tab}}^{(i)}$ and $\Phi_{\text{text}}^{(i)}$ into two risk scores and combines them using Noisy-OR:
\vspace{-4pt}
\begin{equation}
d_{\text{final}}^{(i)} = 1-\left(1-d_{\text{tab}}^{(i)}\right)\left(1-d_{\text{text}}^{(i)}\right).
\vspace{-4pt}
\end{equation}
Finally, $d_{\text{final}}^{(i)}$ is compared with the threshold $\tau$ to determine whether the next step $s_{i+1}$ should be generated by the LRM or the SRM.

\vspace{-4pt}

\subsection{Discussion}
\vspace{-3pt}
\paragraph{Suffix replacement provides cleaner supervision.}
Trajectory-level labeling assigns the same positive label to all steps in an incorrect trajectory, even though many early steps may still be handled correctly by the SRM. This introduces label noise and does not align well with next-step routing. In contrast, suffix replacement identifies a critical routing boundary and assigns the positive label only to the step immediately before the shortest suffix whose SRM replacement causes a stable outcome flip. This yields cleaner step-level supervision and requires no additional human annotation.

\vspace{-3pt}

\paragraph{Offline construction introduces no online overhead.}
Suffix replacement and risk-mapping fitting are performed only once on the held-out validation set. They are fully offline and do not participate in online inference, so they introduce no additional token generation overhead during routing. At test time, EcoTab only computes uncertainty signals and queries the fitted mappings. As shown in the main paper, EcoTab remains lightweight, with routing overhead comparable to STEER and far lower than RSD, SpecCoT, and SpecReason.

\vspace{-3pt}

\paragraph{The fitted mapping generalizes across domains.}
The main paper shows that the learned risk mapping transfers well across domains. In the out-of-domain setting, the mapping fitted on WikiTQ still outperforms Random and STEER when applied directly to TableBench on both Acc@60\% LRM-only FLOPs and FLOPs@98\% LRM-only Acc. Although the in-domain variant performs slightly better, the gap is small, indicating good cross-domain generalization.

\section{Additional Related Work}

\paragraph{Adaptive Reasoning and Efficient Inference.}
The reasoning ability of large language models has been continuously extended through both task-specific adaptation and broader application-oriented studies. Recent surveys have summarized LLM evaluation and benchmarking across diverse tasks and scenarios~\cite{chang2024survey}. Existing work covers optimization and formal reasoning~\cite{liu2026automated,chen2024llms}, knowledge-intensive reasoning~\cite{li-etal-2025-enrich,li2026layerlogitslogicempowering}, software understanding~\cite{zhang2024glitchprober}, and multimodal tasks~\cite{zhang2026temporal,tang2026efficientpostergen,xiao2026staying}. Similar advances have appeared in LLM agents and retrieval-augmented systems, where models interact with external knowledge sources and perform multi-step reasoning~\cite{liu2026toolanchor,shi2026androtmem,yu2026cirrusbench,yu2026lightweight,chen2026learningsearchdecisionbasedagent,wu2026promsa,hao2026evolve,liu2026meta}. To strengthen these capabilities, prior studies employ supervised adaptation, parameter-efficient fine-tuning, reinforcement learning, and knowledge augmentation~\cite{chang2026balora,xiao2026not,liu2026auditingdatamembershipreinforcement,lv2026pcsdpersistentconsistencyselfdistillation,wu2026contrastive,li2026scaling,hao2024towards,hao2026rethinking,wang2026scheduling}. Recent works further investigate adaptive reasoning mechanisms, including structural sensitivity analysis~\cite{he2026order}, step-wise reasoning optimization~\cite{shi2026spader}, and reasoning process manipulation~\cite{liu2025thoughtmanipulationexternalthought,zhou2026look,zhou2026one}.
Alongside capability improvement, reducing inference overhead has become increasingly important. Prior work compresses LLMs through low-bit quantization~\cite{zhao2025quark,zhao2026specquant,xu2026kbvq}, parameter-efficient adaptation and model optimization~\cite{NEURIPS2025_6724eae9}, removes redundant computation in sparse expert architectures~\cite{zhou2025dropping}, or reduces multimodal sequence length through adaptive token compression~\cite{tang2026codeshrink,tang2026efficientpostergen}. These techniques optimize the execution of a given model from the perspectives of parameters, activations, experts, or input tokens. A complementary direction is to avoid assigning identical computational resources to all inference steps. Model routing addresses this issue by leveraging heterogeneous capability--cost trade-offs among models. Stepwise model routing further extends this idea by dynamically allocating models across intermediate reasoning stages, enabling finer-grained control over the capability--efficiency trade-off.

\section{Case Study}
\begin{figure}[h]
    \centering
    \includegraphics[width=0.86\linewidth]{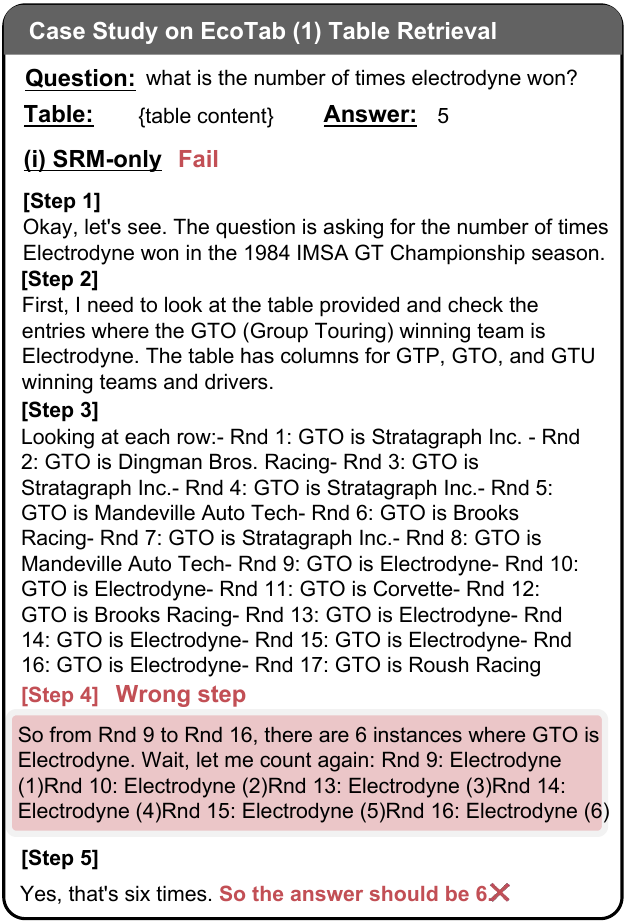}
    \caption{Table Retrieval case with SRM-only.}
    \label{fig:case1}
    \vspace{-10pt}
\end{figure}

\begin{figure}[t]
    \centering
    \includegraphics[width=0.86\linewidth]{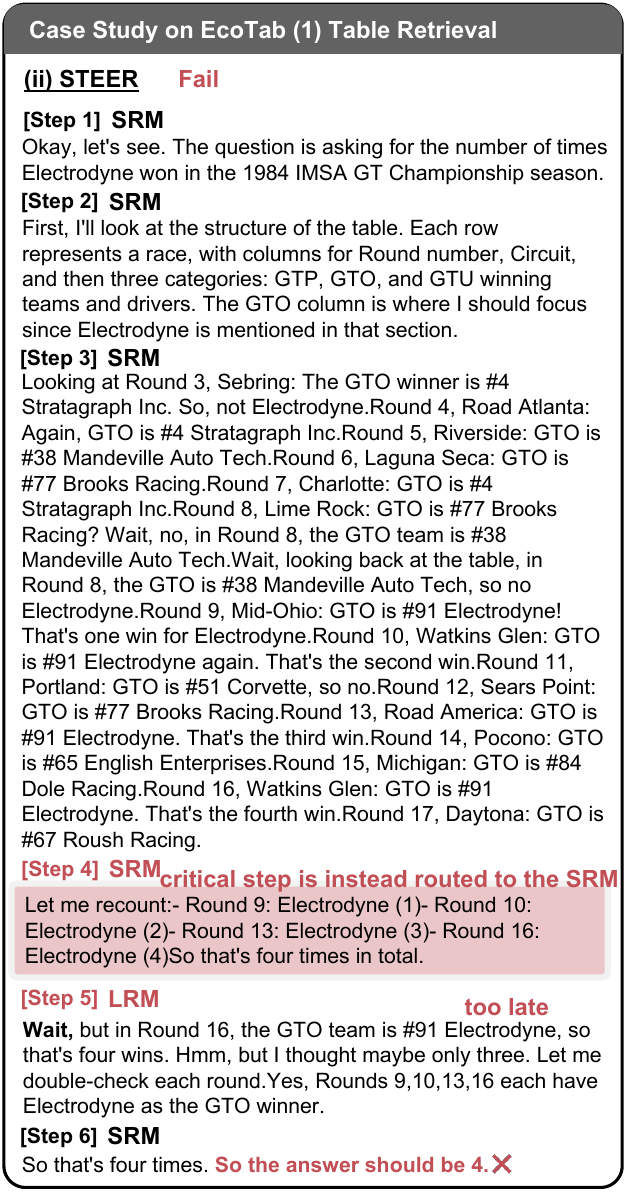}
    \caption{Table Retrieval case with STEER.}
    \label{fig:case2}
    \vspace{-12pt}
\end{figure}

\begin{figure}[t]
    \centering
    \includegraphics[width=0.9\linewidth]{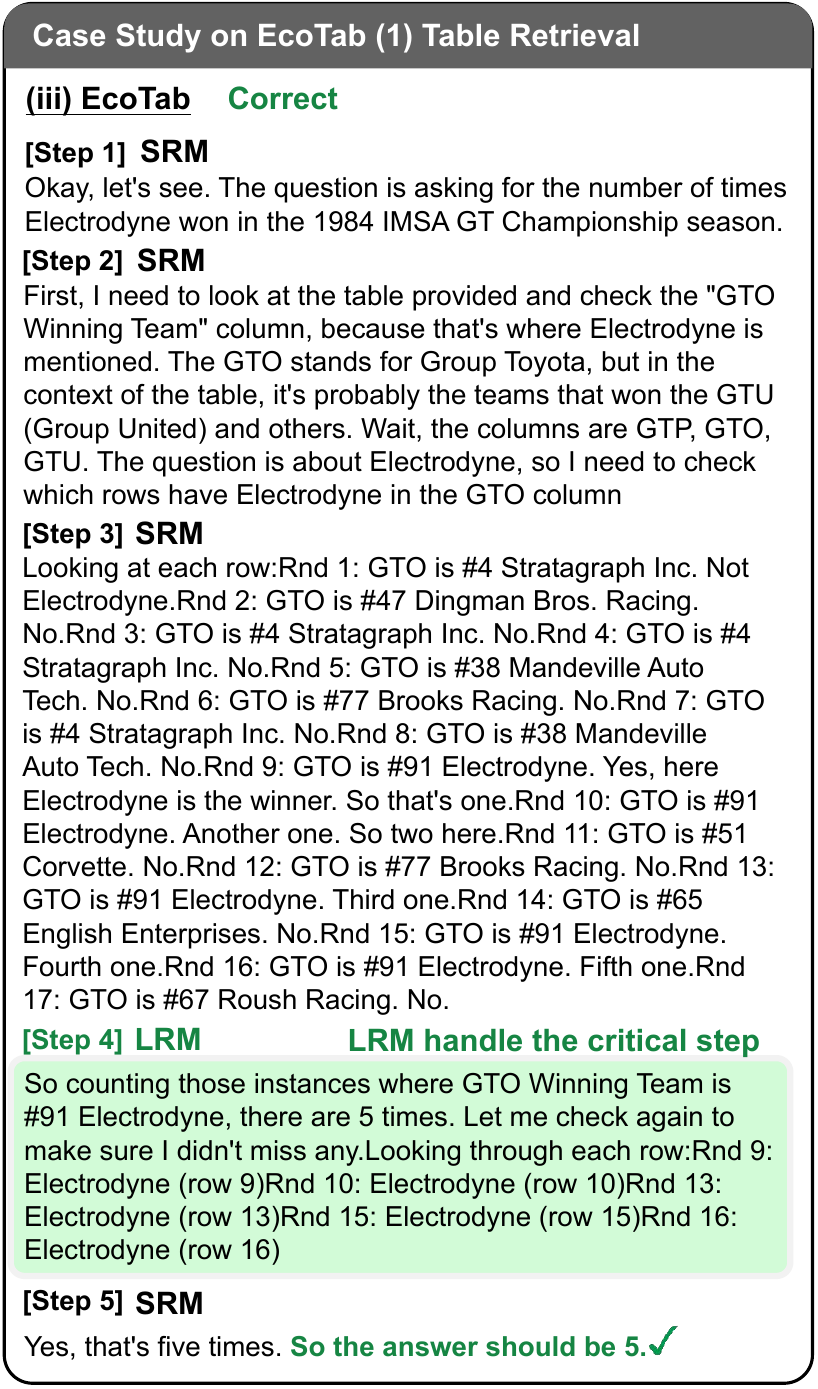}
    \vspace{-4pt}
    \caption{Table Operation case with EcoTab.}
    \label{fig:case3}
    \vspace{-12pt}
\end{figure}

\begin{figure}[h]
    \centering
    \includegraphics[width=0.9\linewidth]{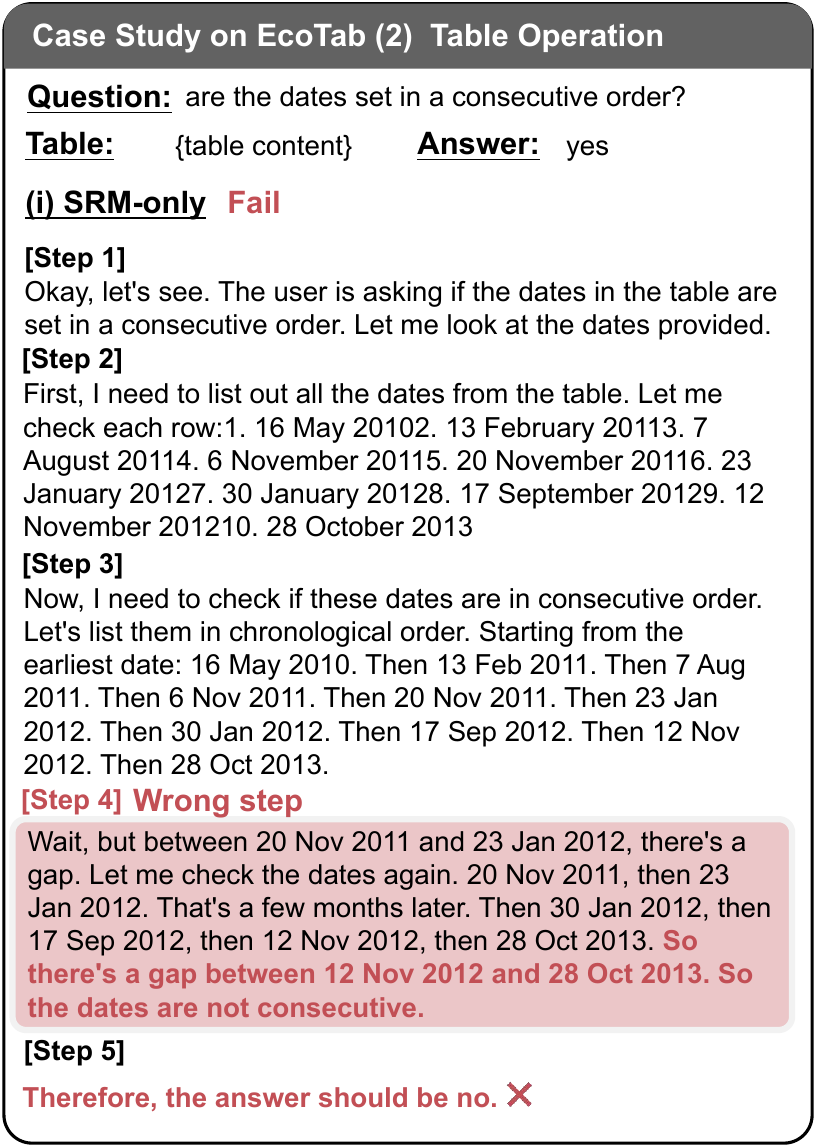}
     \vspace{-4pt}
    \caption{Table Operation case with SRM-only.}
    \label{fig:case4}
    \vspace{-6pt}
\end{figure}

\begin{figure}[t]
    \centering
    \includegraphics[width=0.76\linewidth]{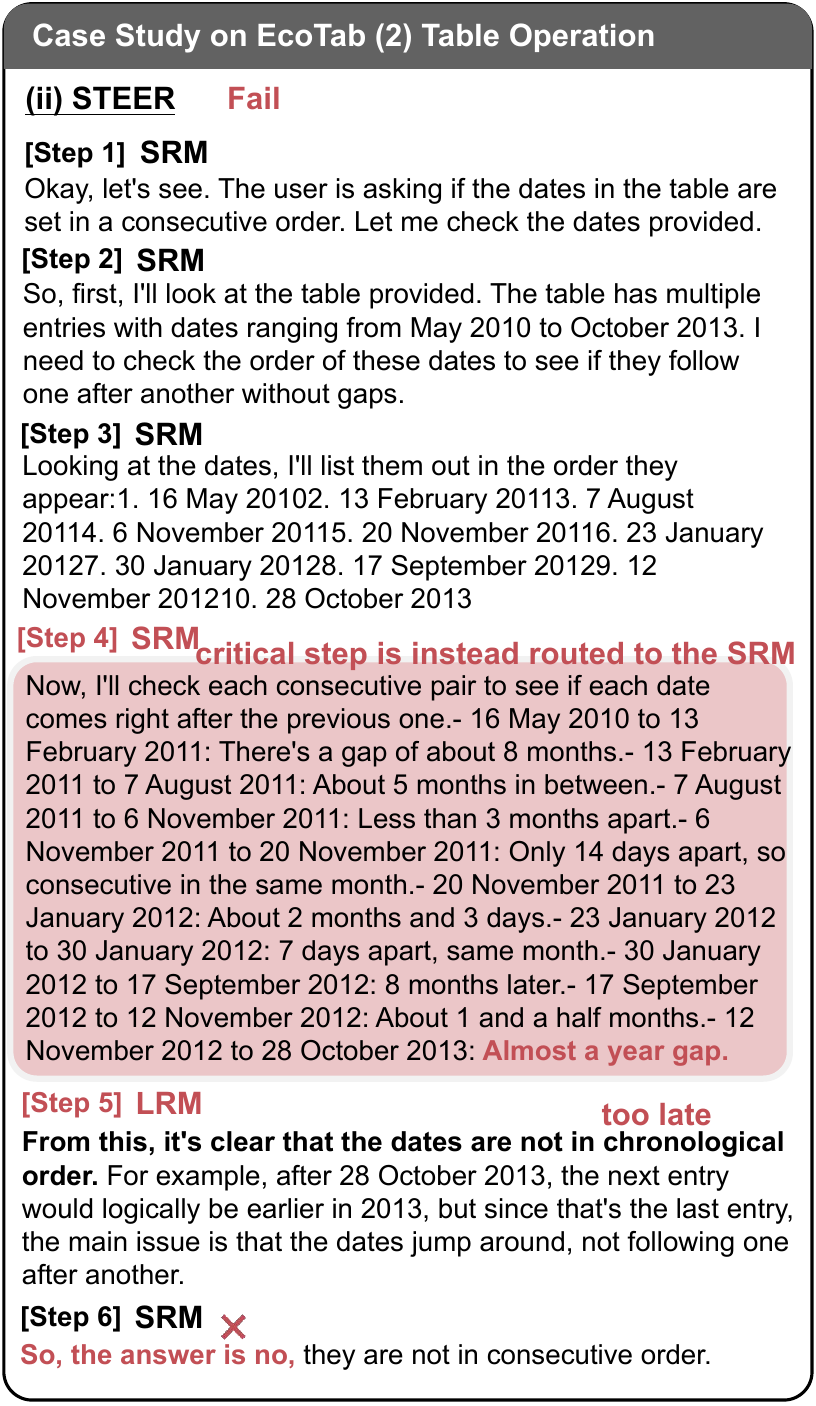}
     \vspace{-4pt}
    \caption{Table Operation case with STEER.}
    \label{fig:case5}
    \vspace{-16pt}
\end{figure}

\begin{figure}[t]
    \centering
    \includegraphics[width=0.76\linewidth]{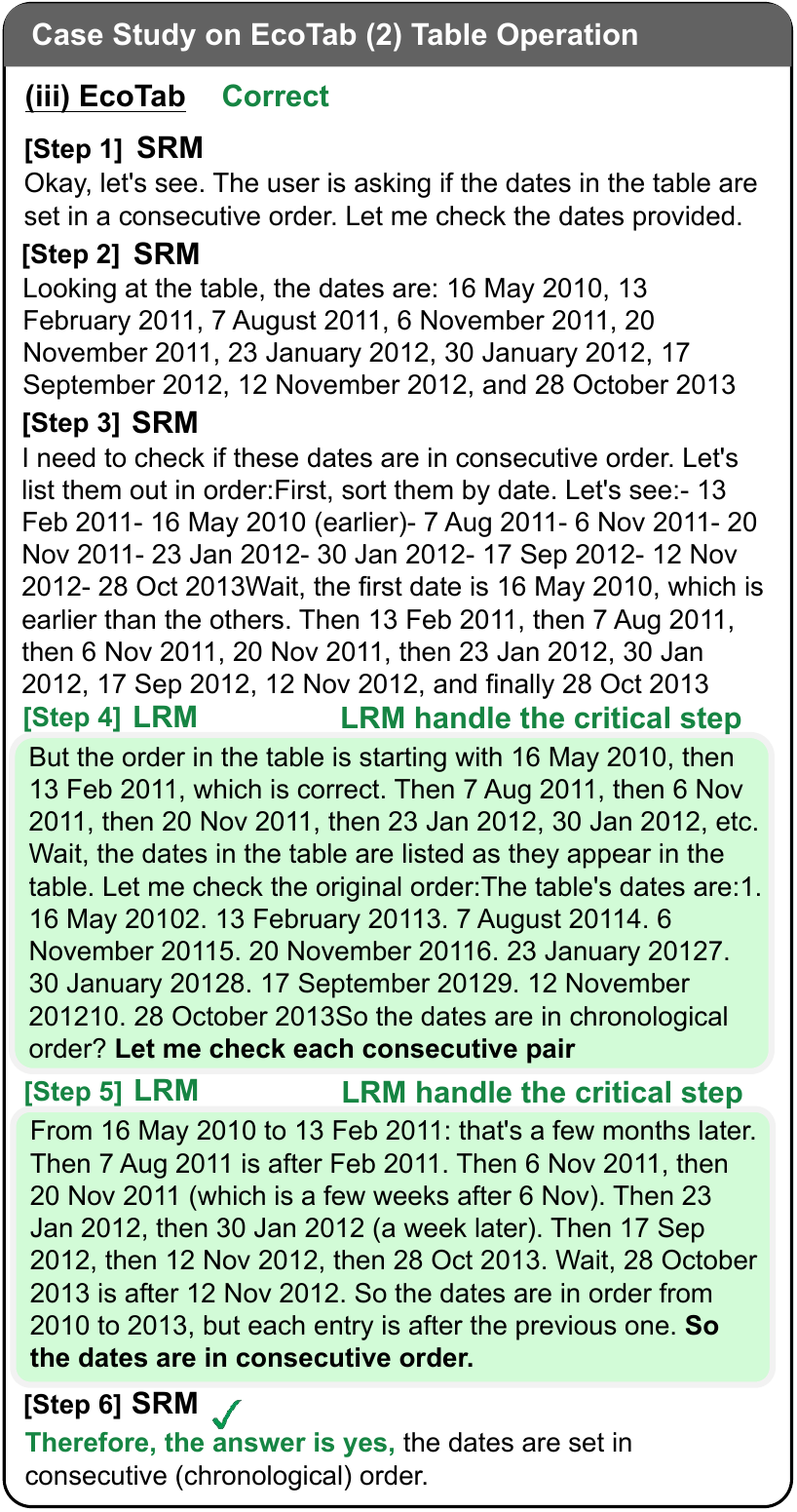}
     \vspace{-4pt}
    \caption{Table Operation case with EcoTab.}
    \label{fig:case6}
    \vspace{-12pt}
\end{figure}

\vspace{-5pt}
We further present representative case studies on two key table reasoning skills, namely Table Retrieval and Table Operation, to qualitatively compare SRM-only, STEER, and EcoTab. Figures~\ref{fig:case1}, \ref{fig:case2}, and \ref{fig:case3} show Table Retrieval cases in which both SRM-only and STEER fail, while EcoTab succeeds. Figures~\ref{fig:case4}, \ref{fig:case5}, and \ref{fig:case6} present analogous cases for Table Operation. The results show that EcoTab can more accurately identify the critical reasoning step and route it to the LRM for handling, leading to correct final predictions.

\end{document}